%% file: main.tex
\documentclass[sigconf]{acmart}
\usepackage{soul}
\usepackage{graphicx}    
\usepackage[utf8]{inputenc}
\usepackage{subfig} 
\usepackage{amsmath}
\usepackage{diagbox}
\usepackage[ruled,vlined]{algorithm2e}
\usepackage{multirow} 
\usepackage{booktabs} 
\usepackage{flushend}

\copyrightyear{2025}
\acmYear{2025}
\setcopyright{acmlicensed}
\acmConference[RACS '25]{International Conference on Research in Adaptive and Convergent Systems}{November 16--19, 2025}{Ho Chi Minh, Vietnam}
\acmBooktitle{International Conference on Research in Adaptive and Convergent Systems (RACS '25), November 16--19, 2025, Ho Chi Minh, Vietnam}
\acmDOI{10.1145/3769002.3769971}
\acmISBN{979-8-4007-2231-8/2025/11}

\acmPrice{15.00}

\begin{document}
\title{Reinforcement Learning-Based Energy-Aware Coverage Path Planning for Precision Agriculture}

\author{Beining Wu}
\affiliation{%
  \institution{EECS Department, South Dakota State University}
  \city{Brookings} 
  \state{South Dakota} 
  \country{USA}
  \postcode{57007}  
}
\email{Wu.Beining@jacks.sdstate.edu}

\author{Zihao Ding}
\affiliation{%
  \institution{EECS Department, South Dakota State University}
  \city{Brookings} 
  \state{South Dakota} 
  \country{USA}
  \postcode{57007}  
}
\email{Zihao.Ding@jacks.sdstate.edu}

\author{Leo Ostigaard}
\affiliation{%
  \institution{EECS Department, South Dakota State University}
  \city{Brookings} 
  \state{South Dakota} 
  \country{USA}
  \postcode{57007}  
}
\email{Leo.Ostigaard@jacks.sdstate.edu}

\author{Jun Huang}
\affiliation{%
  \institution{EECS Department, South Dakota State University}
  \city{Brookings} 
  \state{South Dakota} 
  \country{USA}
  \postcode{57007}  
}
\email{Jun.Huang@sdstate.edu}

\renewcommand{\shortauthors}{B. Wu et al.}

\begin{abstract}
Coverage Path Planning (CPP) is a fundamental capability for agricultural robots; however, existing solutions often overlook energy constraints, resulting in incomplete operations in large-scale or resource-limited environments. This paper proposes an energy-aware CPP framework grounded in Soft Actor-Critic (SAC) reinforcement learning, designed for grid-based environments with obstacles and charging stations. To enable robust and adaptive decision-making under energy limitations, the framework integrates Convolutional Neural Networks (CNNs) for spatial feature extraction and Long Short-Term Memory (LSTM) networks for temporal dynamics. A dedicated reward function is designed to jointly optimize coverage efficiency, energy consumption, and return-to-base constraints. Experimental results demonstrate that the proposed approach consistently achieves over 90\% coverage while ensuring energy safety, outperforming traditional heuristic algorithms such as Rapidly-exploring Random Tree (RRT), Particle Swarm Optimization (PSO), and Ant Colony Optimization (ACO) baselines by 13.4–19.5\% in coverage and reducing constraint violations by 59.9–88.3\%. These findings validate the proposed SAC-based framework as an effective and scalable solution for energy-constrained CPP in agricultural robotics.
\end{abstract}

%
%



\maketitle

\input{section/1_introduction}
\input{section/2_system_model}
\input{section/3_method}
\input{section/4_evaluation}
\input{section/5_result}

\input{section/6_conclusion}

\input{references.bbl}          
\end{document}

%% file: section/1_introduction.tex
\section{Introduction}\label{sec:intro}

Artificial intelligence and automation have greatly advanced modern agriculture by improving productivity and reducing operational costs~\cite{Huang2024TMC,Xing2025ACR}. Agricultural robots are increasingly deployed for tasks such as crop monitoring, soil sampling, automated weeding, and pesticide application~\cite{Raja2024TAE}. A key challenge in their deployment is the Coverage Path Planning (CPP) problem, which involves designing optimal paths that ensure complete area coverage while minimizing resource consumption. Inefficient CPP results in path overlaps, uncovered regions, and excessive energy usage, ultimately reducing system efficiency and lifespan. While much of the existing research emphasizes coverage maximization and task efficiency~\cite{Mier2023RAL,Huang2019TII}, practical deployments are often constrained by limited energy availability, particularly for battery-powered robots operating in large-scale agricultural fields. Thus, incorporating energy-aware strategies into CPP is critical for the real-world viability of autonomous agricultural systems~\cite{Fang2025ARXIV,Fang2026GLOBECOM,Huang2019IWC}.

Existing approaches to energy-constrained CPP fall into three categories: traditional optimization, heuristic algorithms, and learning-based methods. Traditional optimization techniques—such as linear and mixed-integer programming—are effective in structured environments but struggle with scalability in dynamic field conditions~\cite{Wu2025ARXIVb,Ning2019VTM}. For instance, Wang et al.\cite{Wang2024RAL} proposed an artificial potential field-based method for multi-robot coverage in dynamic environments, yet it overlooks energy constraints. Similarly, Fu et al.\cite{Fu2024TIE} introduced a multi-objective framework that considers coverage and trajectory optimization, but does not address energy management. While these methods contribute to geometric efficiency, they fall short in accounting for energy limitations and the necessity of safe return mechanisms, both of which are vital in agricultural applications~\cite{Huang2018COMMAG}.

On top of these foundational efforts, researchers have explored various energy management strategies for robotic systems. Heuristic algorithms, while offering flexibility, often require extensive parameter tuning~\cite{ye2023free,Pan2023SCIS}, limiting their adaptability to dynamic and uncertain environments. Datsko et al.\cite{Datsko2024RAL} introduced a minimum-energy coverage method for UAV swarms by optimizing flight speed and energy consumption, yet did not account for sustained operations via recharging. Lin et al.\cite{Lin2022RAL} proposed a partition-based approach for energy-constrained UAVs supported by mobile charging stations, although their focus remained on point-based surveillance rather than full-area coverage. Despite these advances, energy management and coverage planning are frequently treated as decoupled tasks, overlooking their inherent interdependence in continuous operation scenarios.

Reinforcement Learning (RL) has emerged as a compelling solution due to its capacity for learning adaptive control policies through interaction with the environment~\cite{Wu2025WASA, ye2024multiplexed}. RL has demonstrated effectiveness across diverse domains, including autonomous navigation, robotics, and intelligent decision systems~\cite{wu2023access, wu2023MPE,Fang2022JSAC}. In the context of CPP, Kwon et al.\cite{Kwon2024ICRA} improved energy efficiency through RL-based navigation, albeit focusing on computational rather than physical energy constraints.These efforts, while valuable, tend to treat energy management or coverage efficiency in isolation, limiting their applicability to real-world agricultural systems.

Among RL algorithms, Soft Actor-Critic (SAC) presents unique advantages through its entropy-regularized objective, which promotes effective exploration and enhances policy stability~\cite{Haarnoja2018,DingICNC2025,Fang2025TON}. SAC’s twin Q-network architecture mitigates overestimation bias~\cite{Fujimoto2018}, reducing the risk of unsafe actions that could deplete energy reserves. These characteristics make SAC particularly well-suited for energy-constrained CPP tasks in uncertain, obstacle-rich agricultural environments. Yet, despite its potential, SAC remains underexplored in this domain~\cite{Ding2025IPCCC,Wu2025ToN,Wu2025MNET,Fang2024TMC,Fang2022IOTJ}.

To sum up, existing studies present the following research gaps:  1) coverage algorithms often ignore energy constraints and safe return guarantees; 2) energy management techniques typically operate independently of coverage planning; 3) 
existing RL-based methods lack an integrated framework that jointly addresses coverage efficiency, energy constraints, and operational safety.


To address these gaps, we propose a Soft Actor-Critic-based CPP framework for grid-based agricultural environments with obstacles and charging stations. By embedding energy safety constraints into the learning process, our framework ensures both comprehensive area coverage and mission completion within the available energy budget. The key contributions of this work are as follows:

\begin{itemize}
\item We formally formulate the energy-constrained CPP problem with a return-to-start requirement, incorporating key factors such as coverage efficiency, energy safety, and obstacle avoidance in grid environments. To address this problem, we develop an SAC-based RL approach that leverages convolutional neural networks (CNNs) for spatial feature extraction and long short-term memory (LSTM) networks to account for temporal dependencies.

\item We design a multi-objective reward function that jointly encourages maximum coverage, penalizes excessive energy consumption, and enforces a safe return to the starting point, ensuring energy-aware path planning. Through comprehensive experiments in diverse grid-based environments, we evaluate the proposed approach under varying obstacles and charging station configurations, demonstrating its effectiveness and robustness with over 90\% coverage rates and significant reductions in constraint violations compared to baseline methods.
\end{itemize}

The rest of this paper is organized as follows: Section~\ref{sec:sm} describes the system model, Section~\ref{sec:pfm} presents our design approach, Section~\ref{sec:perfevl} provides experimental results, and Section~\ref{sec:conclusion} concludes the paper.

%% file: section/2_system_model.tex
\section{System Model and Problem Formulation} \label{sec:sm}

\subsection{Environment Modeling}
Consider an agricultural field represented as a planar region $\mathcal{F}$ containing a set of charging stations $\mathcal{C} = \{c_1, c_2, ..., c_m\}$ positioned at strategic locations. The field may contain various agricultural obstacles such as irrigation equipment, storage facilities, or restricted areas, denoted as set $\mathcal{O}$. For computational tractability, $\mathcal{F}$ is discretized into an $N \times N$ grid, where each cell represents a unit area for agricultural operations.

\subsection{Mobile Robot Model}
We consider an autonomous mobile robot $\mathcal{R}$ initially positioned at a base charging station $c_b \in \mathcal{C}$. The robot has physical dimensions of $w \times w$ that fits within a single grid cell. At each discrete time step $t$, $\mathcal{R}$ can execute one of four cardinal movements $\mathcal{A} = \{\text{North}, \text{South}, \text{East}, \text{West}\}$ to adjacent cells, provided they are not occupied by obstacles.

The robot is equipped with:
\begin{itemize}
    \item A localization system (GPS/RTK-GPS) providing precise positioning,
    \item Obstacle detection sensors (LiDAR/proximity sensors) with single-cell lookahead capability,
    \item An energy monitoring system reporting remaining battery level $e(t)$.
\end{itemize}

The robot's energy consumption model follows:
\begin{equation}
    e(t+1) = e(t) - \eta_m d(p_t, p_{t+1}),
\end{equation}
where $e(t)$ represents the energy level in joules (J), $\eta_m$ is the energy consumption rate per unit distance in joules per meter (J/m), and $d(p_t, p_{t+1})$ denotes the distance between consecutive positions $p_t$ and $p_{t+1}$ in meters (m).

\subsection{Task-Specific Constraints}
The model incorporates two essential constraints:
\begin{enumerate}
    \item Coverage Constraint: The robot must visit all accessible cells in $\mathcal{F}$ at least once,
    \item Return Constraint: The robot must return to any charging station $c_i \in \mathcal{C}$ before energy depletion.
\end{enumerate}

\subsection{Safe Cell}
A cell $p \in \mathcal{F}$ is considered safe if it satisfies the following conditions:
\begin{enumerate}
    \item It is not occupied by any obstacle ($p \notin \mathcal{O}$),
    \item There exists at least one feasible path to the nearest charging station with current energy level,
    \item The path consists of consecutive unobstructed cells.
\end{enumerate}

\subsection{State Representation}
The environment state at time $t$ is represented by a 4-channel tensor $\mathcal{S}_t \in \mathbb{R}^{4 \times N \times N}$, where:
\begin{itemize}
    \item Channel 1: Binary obstacle map $\mathbf{O}$,
    \item Channel 2: Charging station locations $\mathbf{C}$,
    \item Channel 3: Current robot position $\mathbf{P}_t$,
    \item Channel 4: Cumulative coverage map $\mathbf{M}_t$.
\end{itemize}

This four-dimensional representation captures all necessary information for the robot to make informed decisions about its next movement while considering both coverage objectives and energy constraints.

\subsection{Problem Formulation}
Based on the aforementioned system model, we formulate the energy-constrained coverage path planning problem in precision agriculture as the following optimization problem:

\begin{equation}
    \begin{aligned}
        &\max_{Q} \sum_{t=1}^T[\psi_1\sum_{n=1}^N C_n(t) - \psi_2E(t)],\\
        \text{s.t.} &
        \begin{cases}
            \text{C1}: e(t) \geq e_{min}, \forall t,\\
            \text{C2}: \cup_{t=1}^T M_t = \mathcal{F}, \\
            \text{C3}: d(p_t, c_i) \leq d_{max}, \forall p_t \in \mathcal{F}, c_i \in \mathcal{C}, \\
            \text{C4}: d(p_t,p_{t+1}) \leq d_{step}, \forall t,
        \end{cases}
    \end{aligned}
\end{equation}


where $Q$ represents the coverage path. $C_n(t)$ represents the coverage reward at time $t$, and $E(t)$ denotes the energy consumption calculated by Eq.(21). $\psi_1$ and $\psi_2$ are weight coefficients. $M_t$ is the set of covered cells at time $t$. $e(t)$ represents the remaining energy at time $t$, and $e_{min}$ is the minimum energy required to return to the nearest charging station. $d(p_t, c_i)$ denotes the Manhattan distance between the current position $p_t$ and charging station $c_i$. $d_{max}$ is the maximum allowed distance to charging stations. $d_{step}$ limits the maximum movement distance between consecutive steps.

Constraint C1 ensures that the robot maintains sufficient energy to return to a charging station. C2 guarantees complete coverage of the field $\mathcal{F}$. C3 ensures the robot stays within reachable range of charging stations. C4 restricts the robot's movement between adjacent cells.

This optimization problem aims to find an optimal path that maximizes the coverage while minimizing energy consumption, subject to energy and movement constraints. The problem is \emph{NP-hard} due to the combination of complete coverage requirements and energy constraints.

%% file: section/3_method.tex
\section{The Proposed Method}\label{sec:pfm}

\subsection{Reinforcement Learning Framework for Energy-Constrained CPP}

\subsubsection{MDP Formulation}
The energy-constrained Coverage Path Planning (CPP) problem is modeled as a Markov Decision Process (MDP), defined as $\mathcal{M} = (\mathcal{S}, \mathcal{A}, \mathcal{P}, \mathcal{R}, \gamma)$, where:

\begin{itemize}
    \item \textbf{State Space ($\mathcal{S}$)}: The state at time $t$ is represented by a 4-channel tensor $\mathcal{S}_t \in \mathbb{R}^{4 \times N \times N}$:
    \begin{itemize}
        \item \textbf{Channel 1 (Obstacle Map)}: Binary map where $1$ represents obstacles and $0$ represents free space.
        \item \textbf{Channel 2 (Charging Station Map)}: Binary map where $1$ represents charging station locations and $0$ represents other cells.
        \item \textbf{Channel 3 (Current Position)}: Binary map with $1$ at the robot’s current position and $0$ elsewhere.
        \item \textbf{Channel 4 (Coverage Map)}: Binary map where $1$ represents visited cells and $0$ represents unvisited cells.
    \end{itemize}

    \item \textbf{Action Space ($\mathcal{A}$)}: The robot can take one of four discrete actions:\\ $$\mathcal{A} = \{\text{Up}, \text{Down}, \text{Left}, \text{Right}\}.$$

    \item \textbf{Reward Function ($\mathcal{R}$)}: The reward at each step balances coverage, energy efficiency, and safety:
    \begin{equation}
        R(s_t, a_t) = \psi_1 \cdot C(s_{t+1}) - \psi_2 \cdot E(a_t) + \psi_3 \cdot S(s_{t+1}),
        \label{eq:reward}
    \end{equation}
    where:
    \begin{itemize}
        \item $C(s_{t+1})$: Reward for visiting a new cell ($+1$) or penalty for revisiting ($-1$).
        \item $E(a_t)$: Energy cost of the action.
        \item $S(s_{t+1})$: Penalty ($-3$) if the robot lacks sufficient energy to reach a charging station, or a small reward ($+0.5$) if it does.
        \item $\psi_1$,$\psi_2$,$\psi_3$ are the weights of the respective reward factors. These weights control the trade-off between coverage maximization ($\psi_1$), energy conservation ($\psi_2$), and safety enforcement ($\psi_3$), where higher values emphasize the corresponding objective. The configuration $\psi_1=1.0$, $\psi_2=0.5$, $\psi_3=0.7$ was determined through sensitivity analysis to achieve optimal balance between coverage efficiency and constraint satisfaction, as demonstrated in Table~\ref{tab:weights}.
    \end{itemize}
\end{itemize}

\subsubsection{Network Architecture}
\begin{figure}[t]
    \centering
    \includegraphics[width=0.9\linewidth]{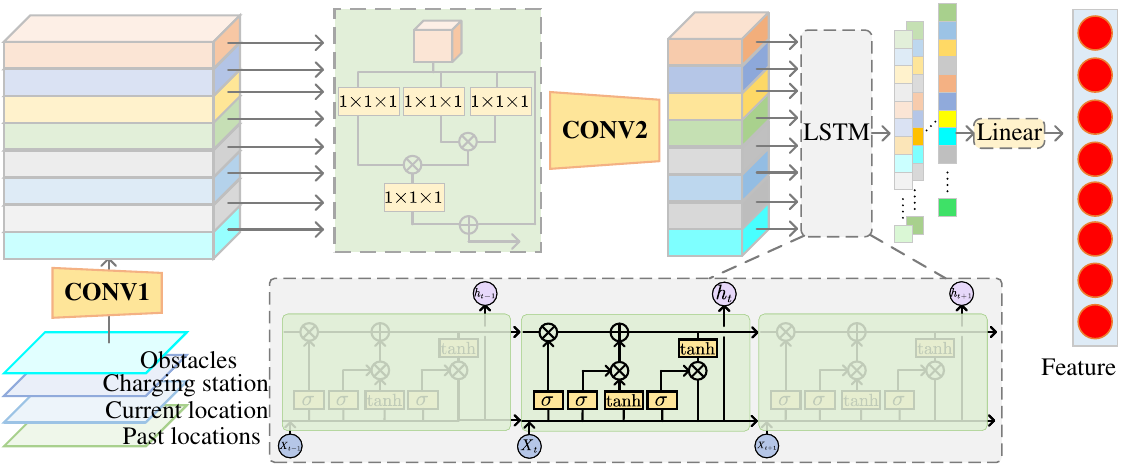}
    \caption{Attention-augmented dual-convolutional architecture for energy-constrained CPP.}
    \label{fig:network}
\end{figure}

As shown in Figure~\ref{fig:network}, the proposed network architecture consists of a hierarchical structure that combines convolutional processing, self-attention mechanisms, and sequential learning. The network processes the 4-channel input tensor through several specialized modules:

\begin{itemize}
    \item \textbf{First Convolutional Layer (CONV1):} As illustrated in Figure~\ref{fig:network}, the initial processing stage applies convolutional operations to extract low-level spatial features from the 4-channel input tensor (obstacle map, charging station locations, current position, and coverage history).
    
    \item \textbf{Multi-Head Self-Attention (MHSA):} Following the first convolutional layer, the MHSA module processes the feature maps to capture global spatial dependencies. This mechanism allows the network to:
    \begin{itemize}
        \item Model long-range dependencies across the environment
        \item Integrate information from charging stations and obstacles effectively
    \end{itemize}
    
    \item \textbf{Second Convolutional Layer (CONV2):} After the attention mechanism, a second convolutional layer further refines the feature representation, focusing on local spatial patterns and their relationships \cite{setitra2023Network}.
    
    \item \textbf{LSTM Layer :} The processed spatial features are then fed into an LSTM layer, which:
    \begin{itemize}
        \item Captures temporal dependencies in the robot's movement sequence
        \item Maintains historical context for decision-making
        \item Enables long-term planning capabilities
    \end{itemize}
    
    \item \textbf{Linear Output Layer:} As depicted in Figure~\ref{fig:network}, the final fully connected layer maps the processed features to:
    \begin{itemize}
        \item Action probabilities $\pi(a_t | s_t)$ for the policy network
        \item Q-values $Q(s_t, a_t)$ for the critic network
    \end{itemize}
\end{itemize}

This hierarchical architecture effectively combines local feature extraction (through convolutions), global relationship modeling (through self-attention), and temporal dependency processing (through LSTM) to address the complexities of energy-constrained coverage path planning. The sequential arrangement of these components ensures comprehensive processing of both spatial and temporal aspects of the navigation task.

\subsubsection{Training Process}
\begin{algorithm}[t]
\caption{Proposed Algorithm for Energy-Constrained Coverage Path Planning} \label{SAC}
Initialize policy network $\pi_\theta$, Q-networks $Q_{\phi_1}$, $Q_{\phi_2}$, and target Q-networks $Q_{\phi'_1}$, $Q_{\phi'_2}$\;
Initialize replay buffer $\mathcal{D}$\;
\For{each episode}{
    Reset environment and obtain initial state $s_0$\;
    \For{each timestep $t$}{
        Sample action $a_t \sim \pi_\theta(a_t | s_t)$\;
        Execute action $a_t$, observe reward $r_t$ and next state $s_{t+1}$\;
        Store transition $(s_t, a_t, r_t, s_{t+1})$ in $\mathcal{D}$\;
        Sample minibatch $(s, a, r, s') \sim \mathcal{D}$\;
        Compute target $y_t$ using Equation~\ref{eq:target}\;
        Update Q-networks by minimizing Equation~\ref{eq:q_loss}\;
        Update policy network by minimizing Equation~\ref{eq:policy_loss}\;
        Adjust $\alpha$ by minimizing Equation~\ref{eq:entropy_loss}\;
        \If{$t \mod \text{target\_update\_freq} == 0$}{
            Update target networks using Equation~\ref{eq:target_update}\;
        }
    }
}
\end{algorithm}
The training process is based on the Soft Actor-Critic (SAC) algorithm, adapted for discrete action spaces. The objective is to optimize the policy and value networks to maximize cumulative rewards while maintaining entropy regularization. The framework of SAC is illustrated in Algorithm~\ref{SAC}. The key components of the training process are as follows:

The Q-network is updated by minimizing the loss:
\begin{equation}
    \mathcal{L}_Q = \mathbb{E}_{(s_t, a_t, r_t, s_{t+1}) \sim \mathcal{D}} \left[ \left( Q(s_t, a_t) - y_t \right)^2 \right],
    \label{eq:q_loss}
\end{equation}
where the target value $y_t$ is computed as:
\begin{equation}
    y_t = r_t + \gamma \cdot \left( \min_{i=1,2} Q_i(s_{t+1}, a') - \alpha \log \pi(a' | s_{t+1}) \right),
    \label{eq:target}
\end{equation}
and $a' \sim \pi(a | s_{t+1})$.

The policy network is updated by minimizing:
\begin{equation}
    \mathcal{L}_\pi = \mathbb{E}_{s_t \sim \mathcal{D}} \left[ \alpha \log \pi(a_t | s_t) - Q(s_t, a_t) \right],
    \label{eq:policy_loss}
\end{equation}
where $\alpha$ is dynamically adjusted to maintain a target entropy:
\begin{equation}
    \mathcal{L}_\alpha = \mathbb{E}_{a_t \sim \pi} \left[ -\alpha \left( \log \pi(a_t | s_t) + \mathcal{H} \right) \right].
    \label{eq:entropy_loss}
\end{equation}

The target networks are updated periodically:
\begin{equation}
    \phi' \leftarrow \tau \phi + (1 - \tau) \phi',
    \label{eq:target_update}
\end{equation}
where $\tau$ is the soft update rate.

\subsubsection{Algorithm Example: Step-by-Step Calculation}
To illustrate the operation of our proposed algorithm, we present a simplified example with a $5 \times 5$ grid environment. For clarity, we focus on a single training step and show the calculation process in detail.

\textbf{Environment Setup:}
We have a $5 \times 5$ grid with obstacles located at coordinates $1,2$, $3,3$, and $4,1$. A charging station is positioned at $0,0$, which is also where the robot begins. The robot starts with maximum energy $E_{\text{max}} = 20$ units and consumes 1 unit per step. Our reward function uses weights $\psi_1 = 1.0$ for coverage, $\psi_2 = 0.5$ for energy efficiency, and $\psi_3 = 0.7$ for safety concerns.

\textbf{Step 1: Current State Representation}
The state $s_t$ at timestep $t=3$ is represented by a $4$-channel tensor. The first channel shows the obstacle map as a binary matrix with values of 1 marking obstacle positions. The second channel indicates the charging station at position $0,0$. The third channel marks the current robot position, which is $2,1$. The fourth channel maintains the coverage map, showing that positions $0,0$, $1,0$, $1,1$, and $2,1$ have already been visited.

\textbf{Step 2: Action Selection}
Using the current policy $\pi_\theta$, we compute action probabilities: $\pi_\theta(\text{Up}|s_t) = 0.15$, $\pi_\theta(\text{Down}|s_t) = 0.25$, $\pi_\theta(\text{Left}|s_t) = 0.30$, and $\pi_\theta(\text{Right}|s_t) = 0.30$. Based on these probabilities, we sample action $a_t = \text{Right}$.

\textbf{Step 3: Action Execution and Reward Calculation}
After executing the Right action, the robot moves to position $3,1$. This consumes 1 unit of energy, leaving the robot with $E_{t+1} = 17$ units remaining. The robot receives a coverage reward of $+1$ for visiting a new cell, and an energy penalty of $-0.5$ for the energy consumption. A safety check confirms that a path to the charging station exists with the remaining energy, resulting in a safety reward of $+0.5$. The total reward is calculated as $r_t = 1.0 \times 1 - 0.5 \times 1 + 0.7 \times 0.5 = 0.85$.

\textbf{Step 4: Learning Update}
For updating the Q-networks, we compute the target value $y_t$. We first sample an action for the next state, obtaining $a' = \text{Up}$. The Q-values for this state-action pair are $Q_{\phi_1}(s_{t+1}, a') = 1.7$ and $Q_{\phi_2}(s_{t+1}, a') = 1.8$. Taking the minimum value to avoid overestimation, we get $\min_{i=1,2} Q_i(s_{t+1}, a') = 1.7$. The entropy term is calculated as $-\alpha \log \pi(a'|s_{t+1}) = -0.1 \times \log(0.2) = 0.16$, assuming $\alpha = 0.1$. With a discount factor of $\gamma = 0.98$, the target value becomes $y_t = 0.85 + 0.98 \times (1.7 - 0.16) = 2.36$. The Q-networks are then updated using the loss $L_Q = (Q_{\phi_1}(s_t, a_t) - y_t)^2 = (2.1 - 2.36)^2 = 0.0676$, assuming the current Q-value is $2.1$.

\textbf{Step 5: Target Network Update}
If the update frequency condition is met, the target networks are updated using soft updates with the formula $\phi'_1 \leftarrow 0.005 \times \phi_1 + 0.995 \times \phi'_1$, assuming $\tau = 0.005$. This completes one step of the training process. Through multiple episodes, the agent learns to balance coverage efficiency with energy constraints and safety considerations.

%% file: section/4_evaluation.tex
\section{Performance Evaluation} \label{sec:perfevl}

In this section, we validate the effectiveness of the proposed SAC algorithm in the context of coverage path planning through simulations. We compare the performance of SAC against established heuristic algorithms including Rapidly-exploring Random Tree (RRT), Particle Swarm Optimization (PSO), and Ant Colony Optimization (ACO) in terms of trajectory optimization and energy efficiency.

\subsection{Parameter Settings}

\begin{table}[t]
\small
\centering
\caption{Parameters for simulation setup and RL training.}
\label{table:parameters}
\begin{tabular}{l c}
\toprule
\textbf{Parameter} & \textbf{Value} \\
\midrule
Learning Rate for Actor ($\alpha_\text{actor}$) & $3 \times 10^{-4}$ \\
Learning Rate for Critic ($\alpha_\text{critic}$) & $3 \times 10^{-3}$ \\
Learning Rate for Entropy ($\alpha_\text{entropy}$) & $1 \times 10^{-3}$ \\
Discount Factor ($\gamma$) & $0.98$ \\
Replay Buffer Capacity ($|\mathcal{D}|$) & $50,000$ \\
Mini-batch Size & $64$ \\
Target Update Rate ($\tau$) & $0.005$ \\
Exploration Noise Std & $0.1$ \\
Grid Size & $15 \times 15$ \\
Obstacle Density & $10\%-20\%$ \\
Number of Charging Stations & $4$--$6$ \\
Charging Station Distribution & Strategic placement \\
Initial Energy Level ($E_{\text{max}}$) & $100$ units \\
Energy Consumption Rate ($\eta_m$) & $1$ unit/step \\
Minimum Return Energy ($e_{\text{min}}$) & $10$ units \\
Charging Rate & $5$ units/timestep \\
Sensor Range & Single-cell lookahead \\
CNN Kernel Size & $5 \times 5$ \\
CNN Output Channels & $16$ \\
LSTM Hidden Units & $128$ \\
Self-Attention Heads & $4$ \\
Training Hardware & NVIDIA RTX 3090 \\
\bottomrule
\end{tabular}
\end{table}


The simulation settings for the coverage path planning (CPP) system are shown in Table \ref{table:parameters}. The simulations were conducted in Python 3.10 to implement the Deep Neural Network (DNN) in the SAC algorithm. The network structure integrates Multi-Head Self-Attention (MHSA), Convolutional Neural Networks (CNN), Long Short-Term Memory (LSTM), and a linear output layer.

The network architecture consists of multiple modules to process input data efficiently. We employ two convolutional layers, each with a kernel size of 5 and a stride of 1. The first convolutional layer expands from 4 input channels (representing the state) to 16 output channels, followed by a Multi-Head Self-Attention (MHSA) module with 4 attention heads to capture global spatial dependencies. The second convolutional layer further processes the data with 16 output channels, followed by a 2 × 2 max pooling layer for dimensionality reduction. After flattening the output, the data is passed into a Long Short-Term Memory (LSTM) layer with 128 hidden units to capture temporal dependencies. Finally, a linear output layer maps the features to the action probabilities \(\pi(a_t | s_t)\) and Q-values \(Q(s_t, a_t)\).

The Adam optimizer is used for training, with its parameters initialized following a zero-mean normal distribution. The hyperparameters used in the experiments are listed in Table \ref{table:parameters}.

\subsection{Performance Metrics}
To evaluate the effectiveness of our proposed approach, we employ the following quantitative metrics:

\subsubsection{Reward}
The reward metric measures the overall performance, balancing coverage efficiency and energy consumption:
\begin{equation}
R_{\text{total}} = \sum_{t=1}^{T} \left(\psi_1 \cdot C(s_{t+1}) - \psi_2 \cdot E(a_t) + \psi_3 \cdot S(s_{t+1})\right),
\end{equation}
where $C(s_{t+1})$ represents the coverage reward, $E(a_t)$ denotes the energy consumption penalty, and $S(s_{t+1})$ indicates the safety reward at time step $t$. The coefficients $\psi_1$, $\psi_2$, and $\psi_3$ are weighting factors.

\subsubsection{Coverage Rate}
The coverage rate evaluates the proportion of accessible areas successfully covered by the robot:
\begin{equation}
\text{Coverage Rate} = \frac{\sum_{i=1}^{N} \sum_{j=1}^{N} V_{i,j}}{N^2 - |O|} \times 100\%,
\end{equation}
where $V_{i,j}$ is a binary indicator of whether the cell at position $(i,j)$ has been visited, $N^2$ represents the total number of cells in the environment, and $|O|$ denotes the number of cells occupied by obstacles.

\subsubsection{Violations Number}
The violations number refers to the total number of failed episodes throughout the training process, where the agent violates the predefined energy constraints:

\begin{equation}
\text{Violations} = \sum_{t=1}^{T} \mathbb{I}(e_t < e_{\min}),
\end{equation}
where $\mathbb{I}(\cdot)$ is the indicator function that returns 1 when the condition is true and 0 otherwise, $e_t$ is the remaining energy at time step $t$, and $e_{\min}$ represents the minimum energy required to return to the nearest charging station.

\subsubsection{Energy Efficiency}
The energy efficiency metric assesses how effectively the robot utilizes its energy for coverage:
\begin{equation}
\text{Energy Efficiency} = \frac{\sum_{i=1}^{N} \sum_{j=1}^{N} V_{i,j}}{E_{\text{total}}} \times 100\%,
\end{equation}
where $E_{\text{total}}$ is the total energy consumed during the mission.

%% file: section/5_result.tex
\begin{figure*}[ht]
    \centering
    \subfloat[Map 1]{\includegraphics[width=0.2\textwidth]{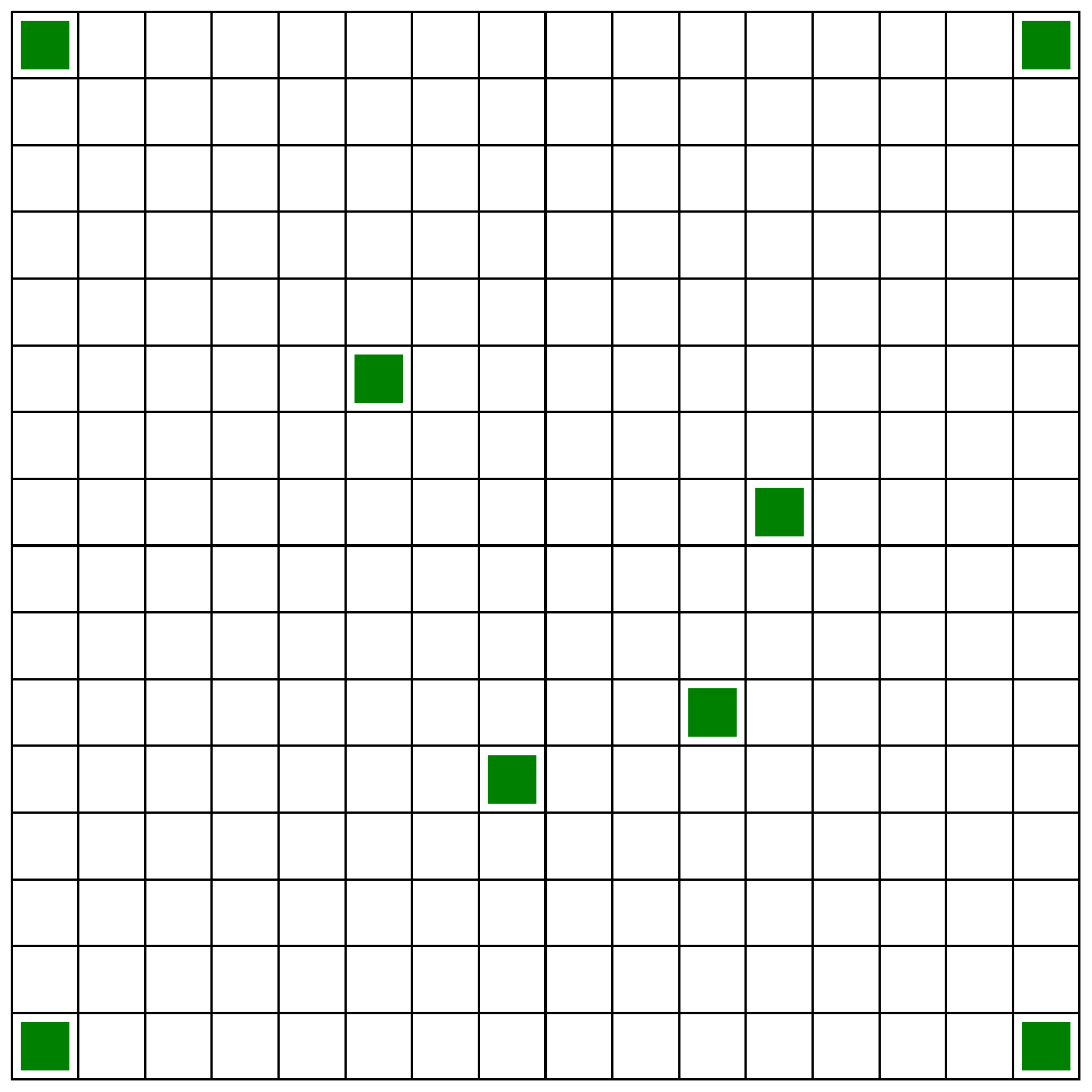}}\hspace{1cm}
    \subfloat[Map 2]{\includegraphics[width=0.2\textwidth]{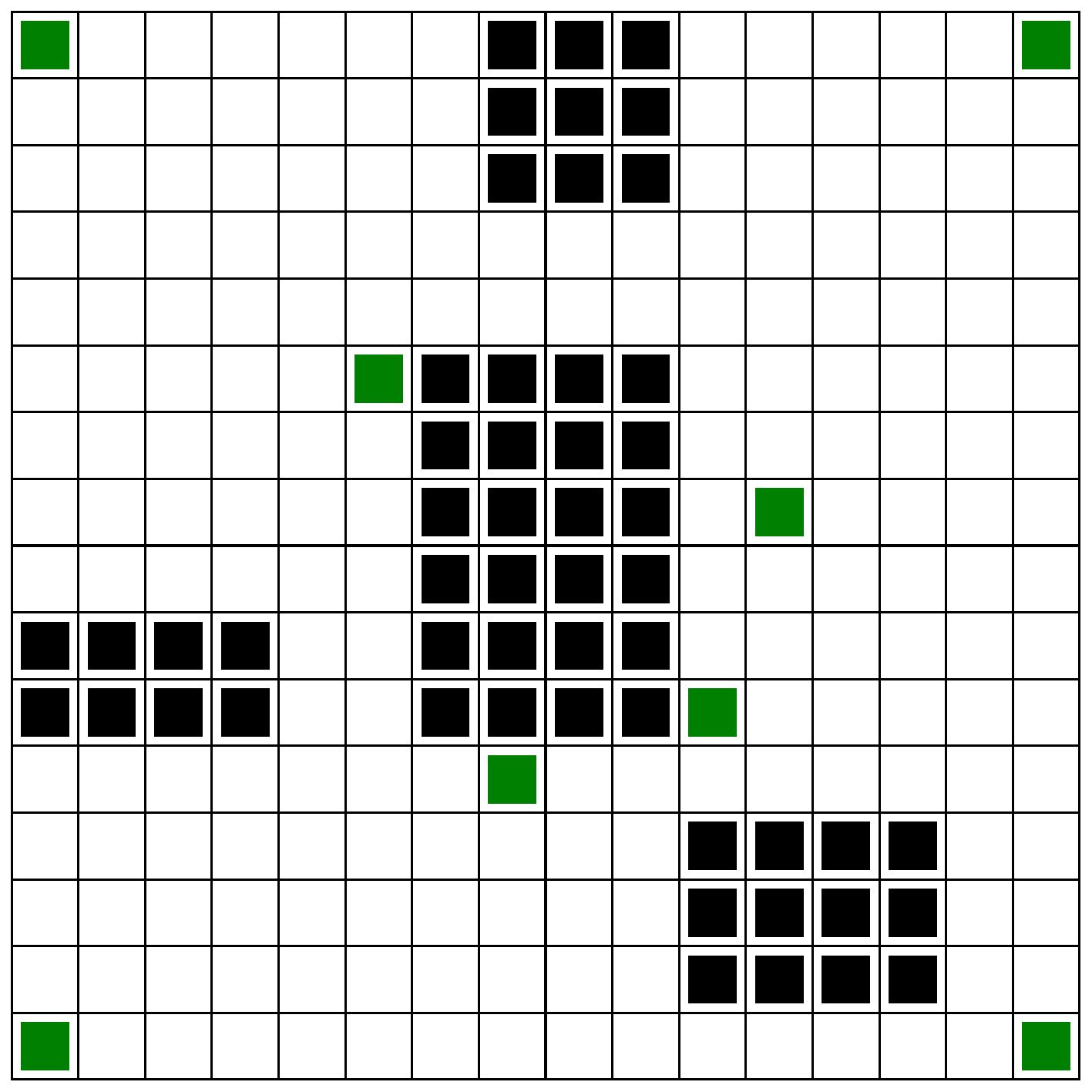}}\hspace{1cm}
    \subfloat[Map 3]{\includegraphics[width=0.2\textwidth]{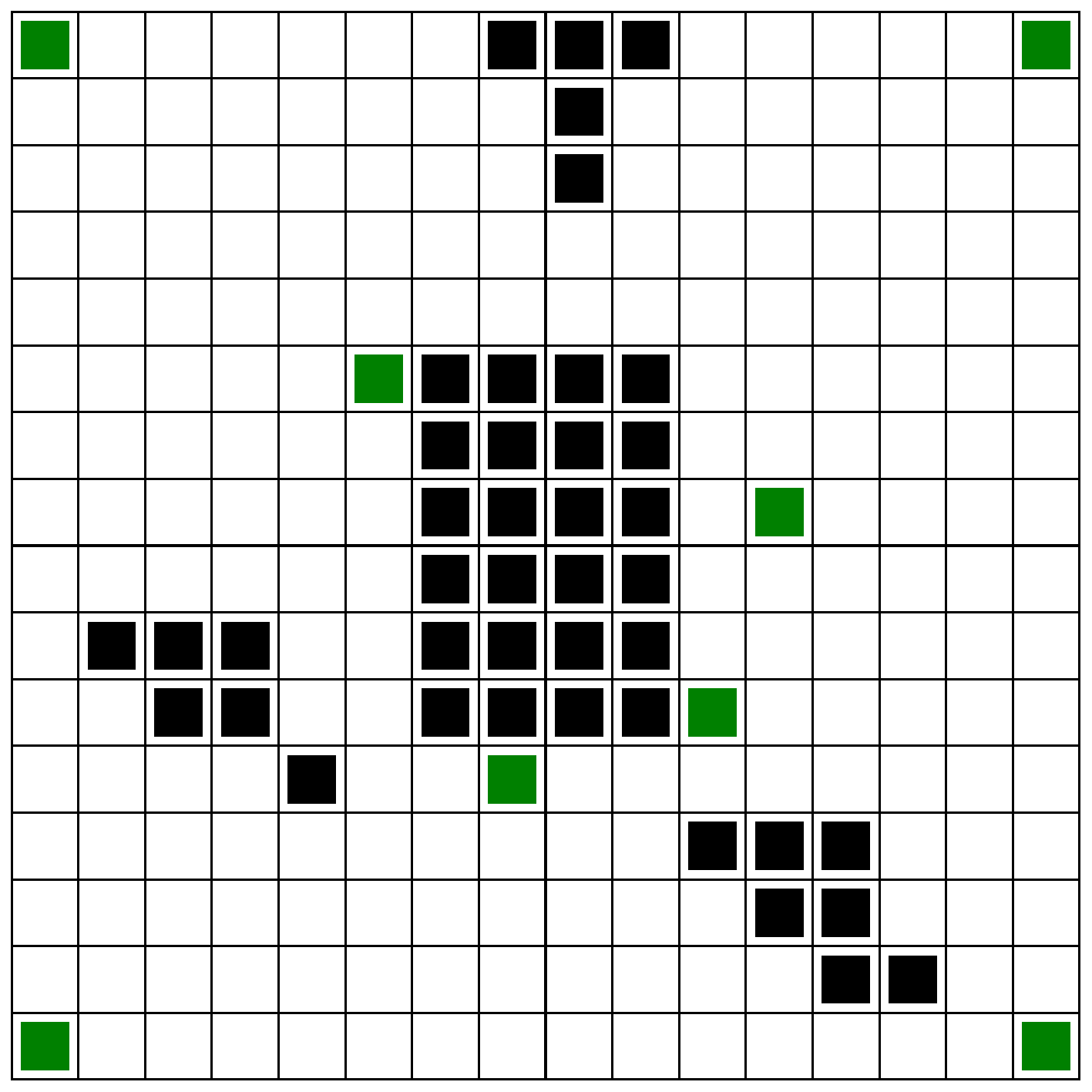}}
    \caption{Three different maps used in our experiments. Green cells indicate the charging stations and the black cells indicate the obstacles. Each cell represents a 1m $\times$ 1m area in the agricultural field. Map 1 (left) shows a simple environment with minimal obstacles and evenly distributed charging stations. Map 2 (center) presents a more complex layout with clustered obstacles representing buildings and equipment. Map 3 (right) features a challenging environment with irregular obstacle patterns and strategic charging station placement.}
    \label{fig:maps}
\end{figure*}

\begin{figure*}[t]
    \centering
    \begin{tabular}{ccc}
        \includegraphics[width=0.3\textwidth]{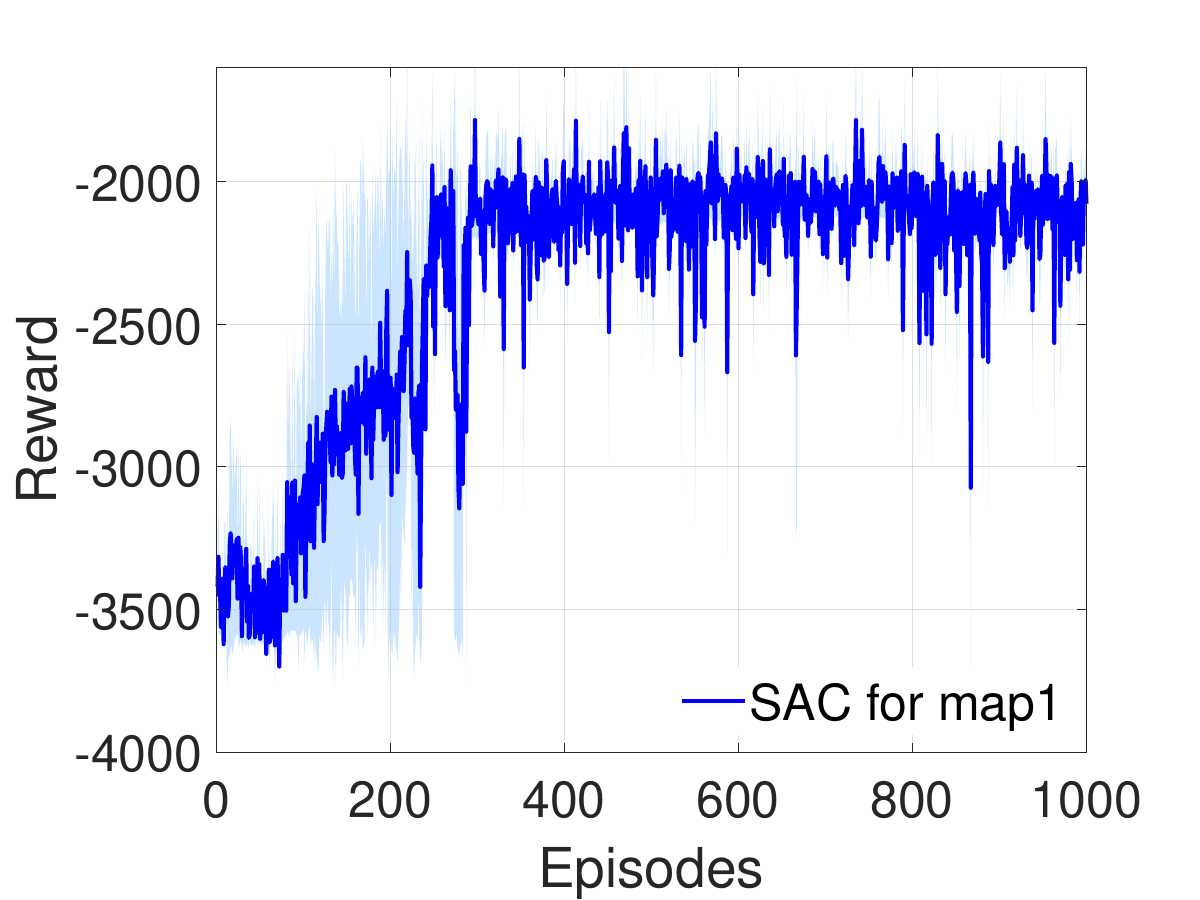} &
        \includegraphics[width=0.3\textwidth]{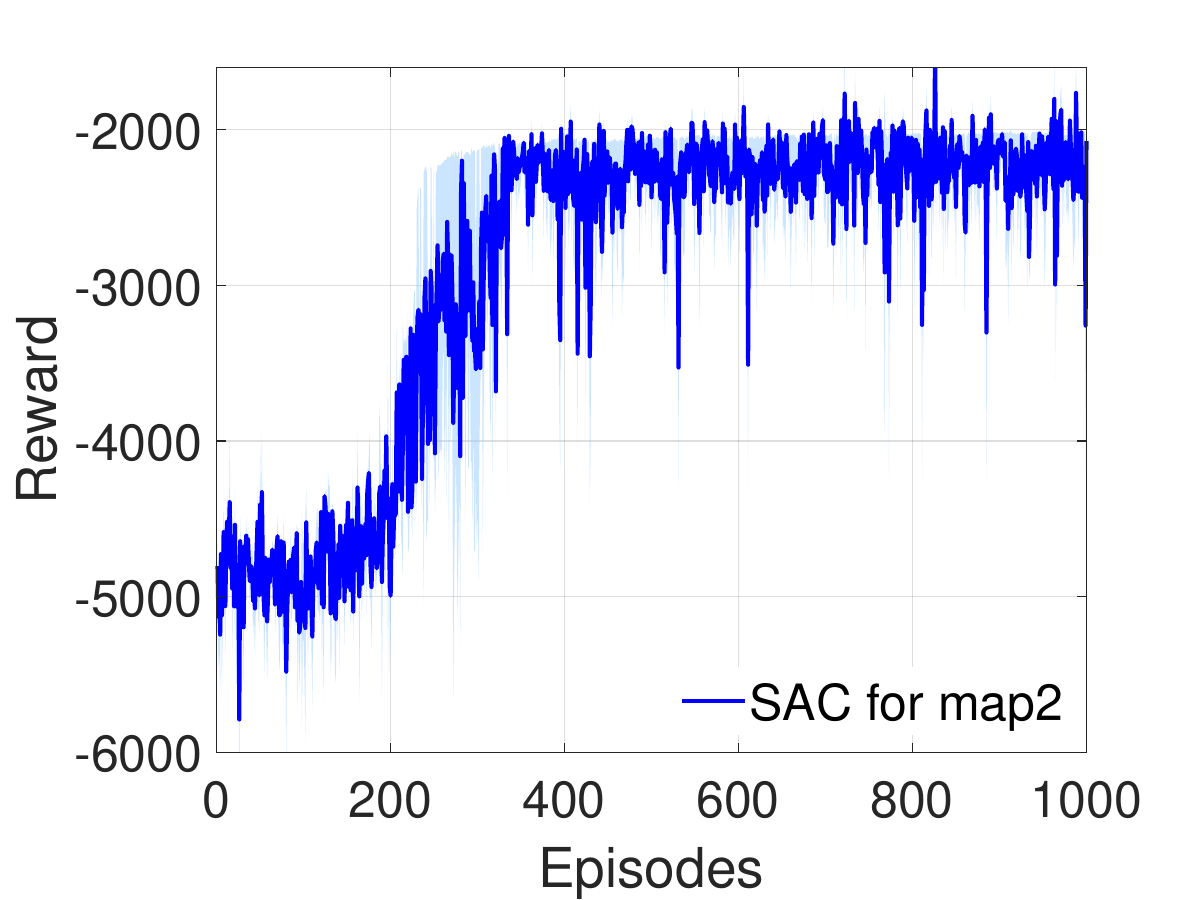} &
        \includegraphics[width=0.3\textwidth]{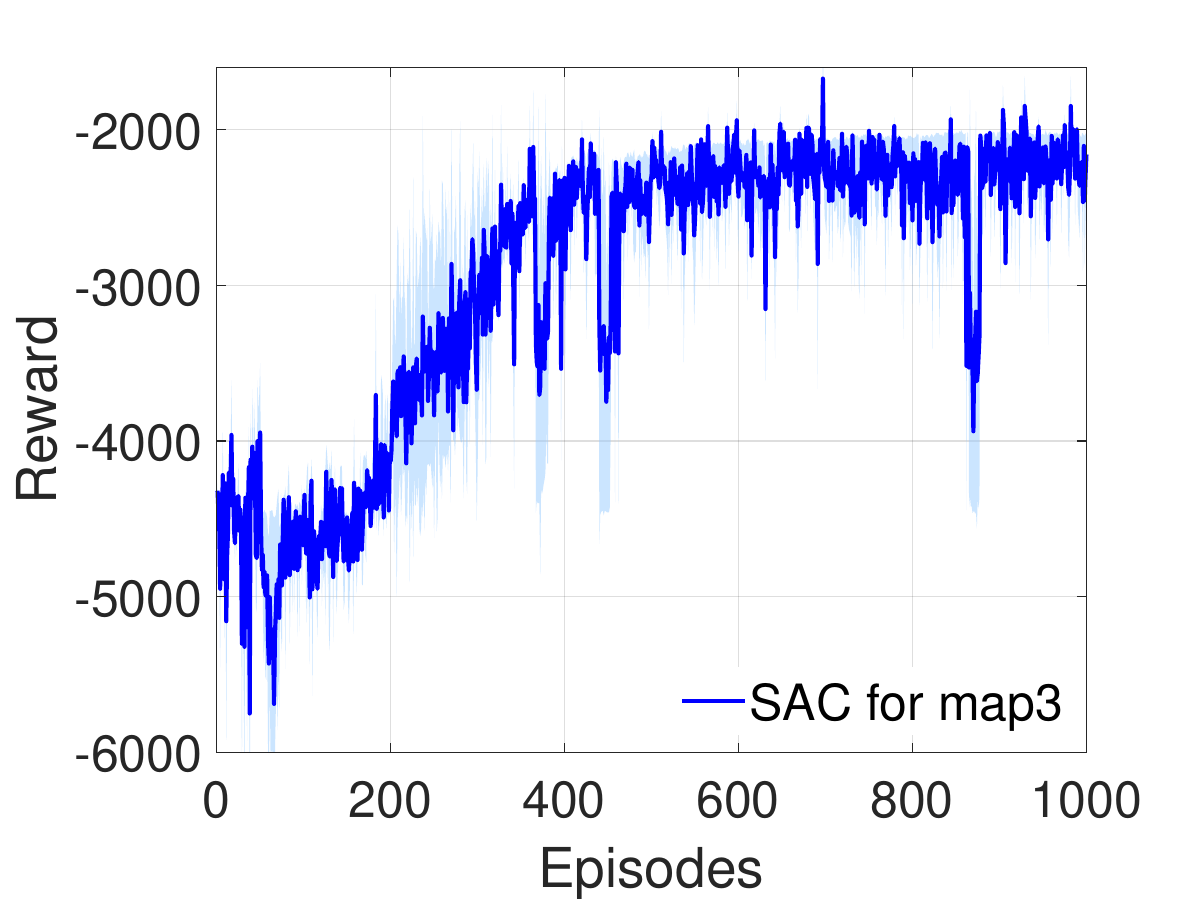} \\
        (a) Reward: Map 1 & (b) Reward: Map 2 & (c) Reward: Map 3
    \end{tabular}
    \caption{Rewards in three training maps. The x-axis represents the number of training episodes, and the y-axis shows the cumulative reward values. Higher reward values indicate better performance in balancing coverage efficiency and energy consumption. Map 1 shows faster convergence due to simpler terrain, while Maps 2 and 3 demonstrate more gradual improvement as the agent learns to navigate more complex environments.}
    \label{fig:reward}
\end{figure*}

\section{Results}
In our experiments, we employed three distinct map configurations, as illustrated in Figure~\ref{fig:maps}, where green cells denote charging stations and black cells indicate obstacles. In practical agricultural environments, the distribution and availability of charging stations, as well as the positions of obstacles, are inherently unpredictable. To systematically evaluate the robustness and adaptability of the proposed algorithm, we selected these three maps as representative test cases, each reflecting different levels of environmental complexity.


The reward variation, as depicted in Figure~\ref{fig:reward}, demonstrates that the SAC algorithm consistently exhibits a stable convergence trend across all three map configurations. Initially, the reward values are low during the early training stages; however, as training progresses, the rewards gradually increase and stabilize after approximately 400 to 600 episodes. Furthermore, as the environmental complexity increases—progressing from Map 1 to Map 2 and Map 3—the convergence rate decelerates. This slowdown is primarily attributed to the greater number of obstacles in Map 2 and Map 3 compared to Map 1, which significantly increases the difficulty of exploration and policy optimization, thereby extending the time required for the algorithm to achieve stable performance.


\begin{figure*}[t]
    \centering
    \begin{tabular}{ccc}
        \includegraphics[width=0.3\textwidth]{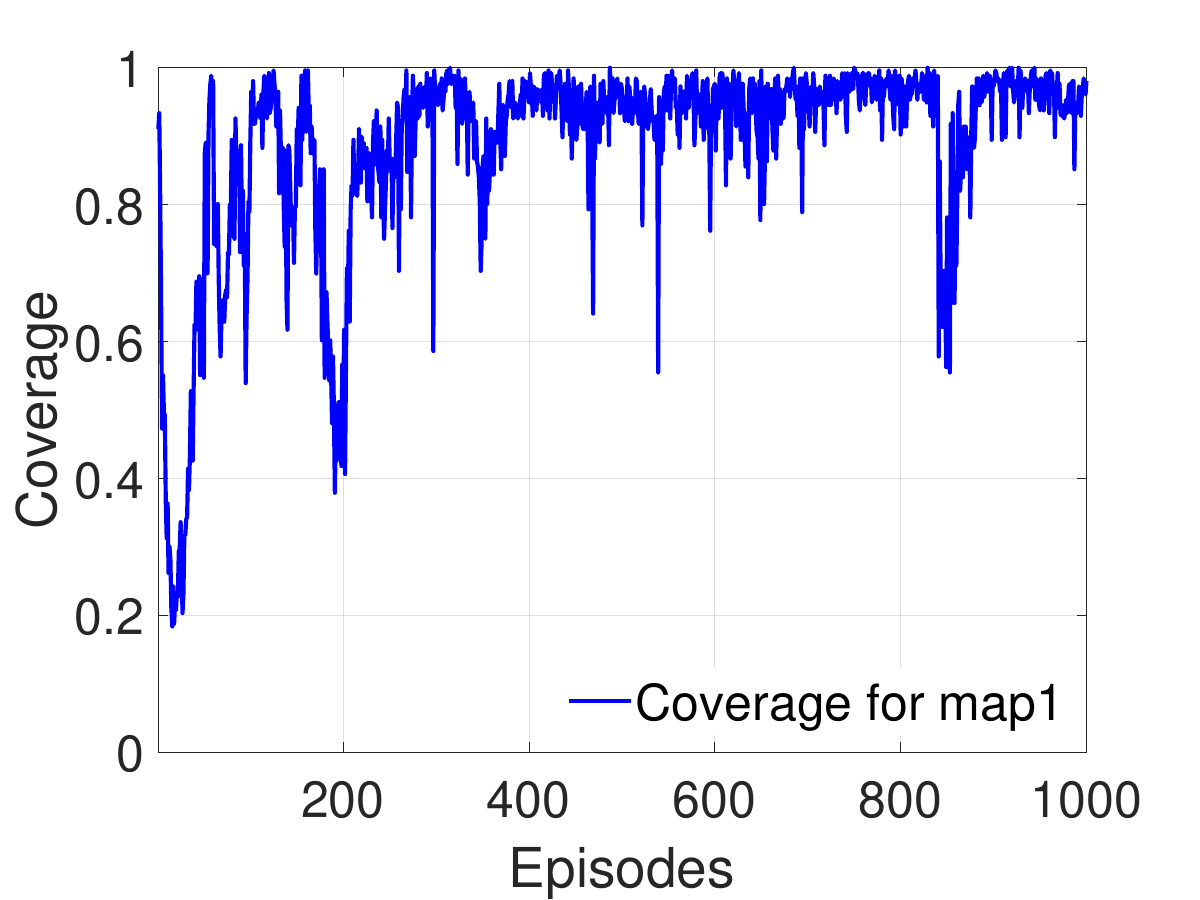} &
        \includegraphics[width=0.3\textwidth]{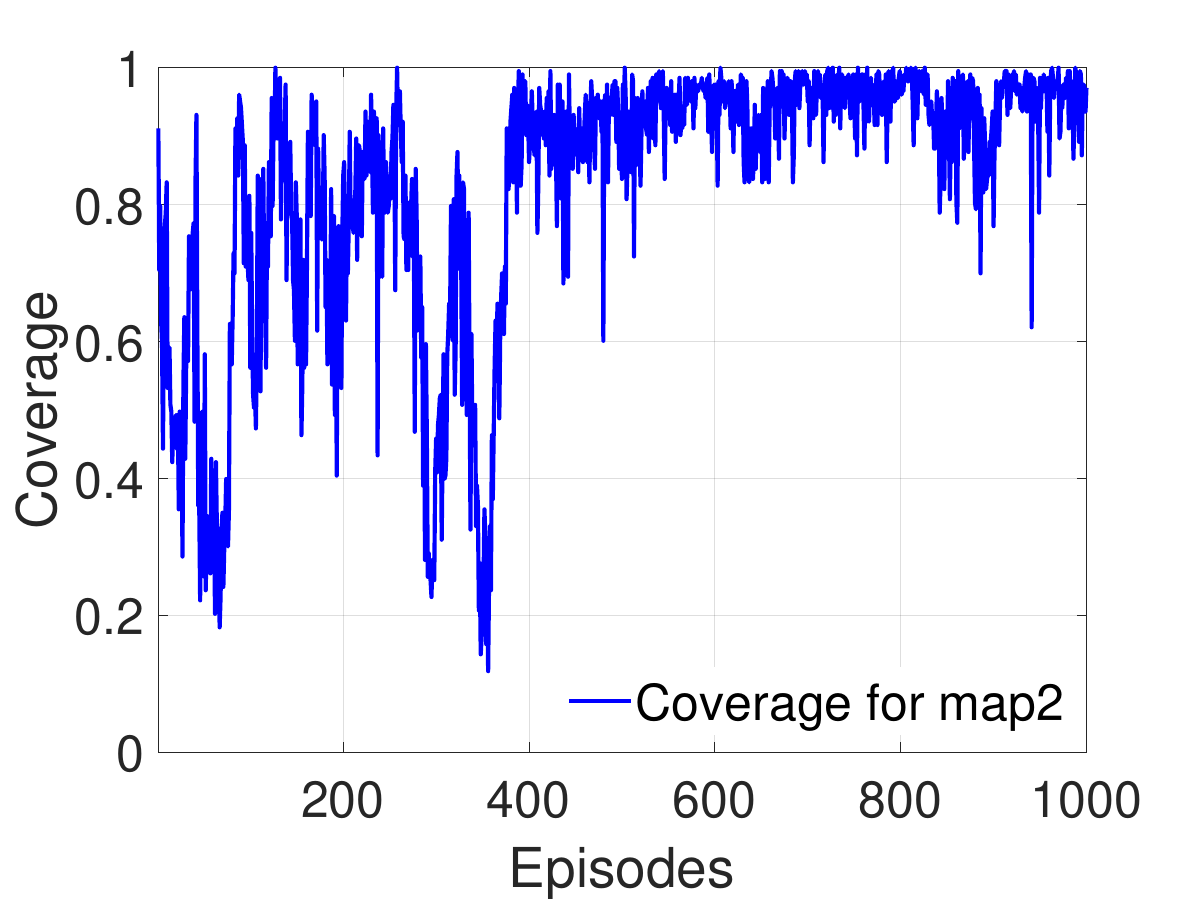} &
        \includegraphics[width=0.3\textwidth]{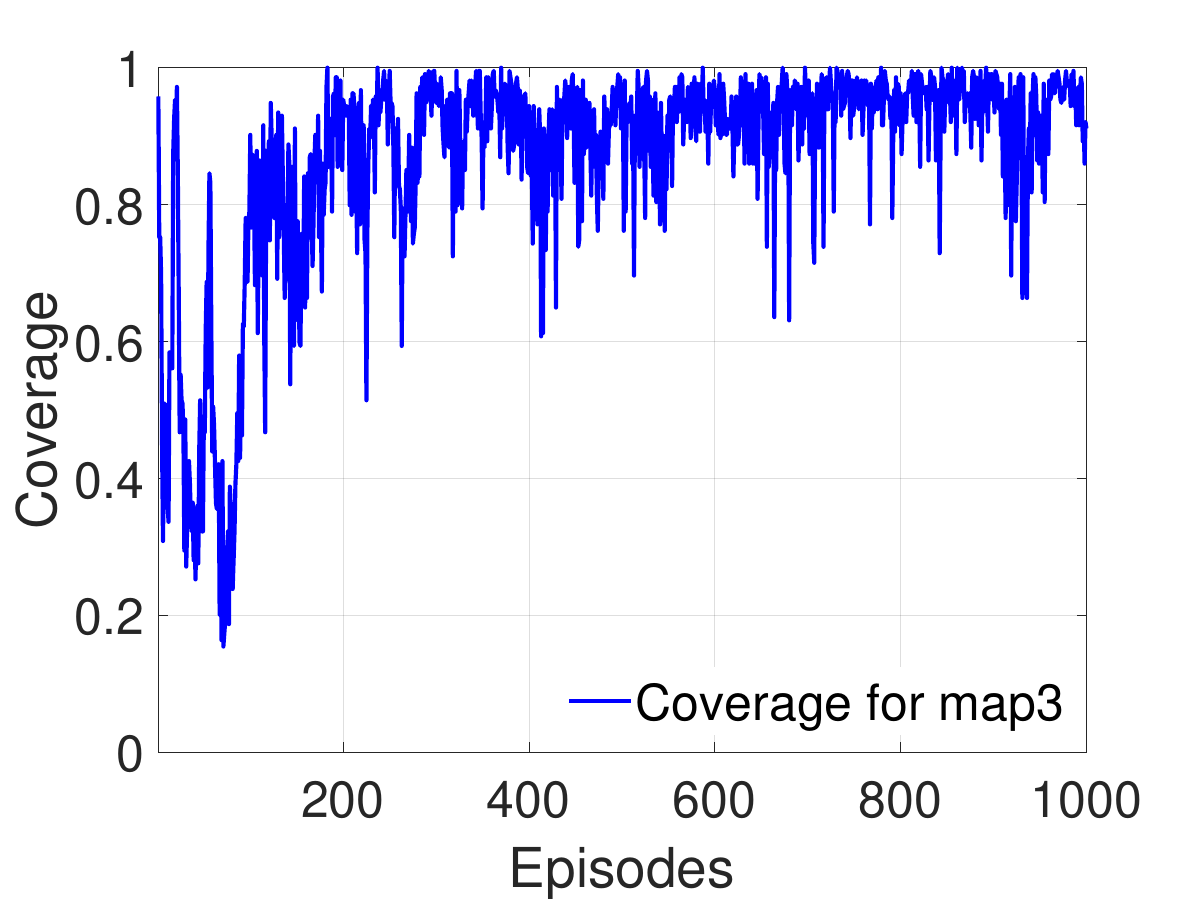} \\
        (a) Coverage: Map 1 & (b) Coverage: Map 2 & (c) Coverage: Map 3
    \end{tabular}
    \caption{Percentages of coverage in three training maps. The x-axis represents the number of training episodes, and the y-axis shows the percentage of accessible areas successfully covered by the robot. Map 1 achieves near-complete coverage (>95\%) after 400 episodes. Maps 2 and 3  show more variability but consistently reach above 90\% coverage in later training stages despite increased environmental complexity.}
    \label{fig:cov}
\end{figure*}

As illustrated in Figure~\ref{fig:cov}, the SAC algorithm achieves high coverage rates across all three map configurations. In Map 1, the coverage rate rapidly converges toward 100\%, indicating efficient path planning in relatively simple environments. In more complex scenarios (Map 2 and Map 3), the coverage rate exhibits greater fluctuations due to the increased number of obstacles; however, it consistently remains above 90\% in the later stages of training. These results highlight the robustness and adaptability of the SAC algorithm, demonstrating its capability to maintain high coverage efficiency even in challenging and obstacle-dense environments.


The reward function in Equation (4) balances coverage efficiency, energy consumption, and safety constraints through weight coefficients $\psi_1$, $\psi_2$, and $\psi_3$. We conducted sensitivity analysis by varying each parameter while fixing others at default values, with 10 independent trials per configuration in Map 1-3.

\begin{table*}[t]
\centering
\caption{Performance with different reward function weight configurations. The \textcolor{red}{red} indicates the best performance, while the \textcolor{blue}{blue} indicates the second-best performance.}
\label{tab:weights}
\begin{tabular}{c c c c c c c}
\toprule
\textbf{Weight Configuration} & \multicolumn{2}{c}{\textbf{Map 1}} & \multicolumn{2}{c}{\textbf{Map 2}} & \multicolumn{2}{c}{\textbf{Map 3}} \\
\cmidrule(lr){2-3} \cmidrule(lr){4-5} \cmidrule(lr){6-7}
 & \textbf{Coverage (\%)} & \textbf{Violations} & \textbf{Coverage (\%)} & \textbf{Violations} & \textbf{Coverage (\%)} & \textbf{Violations} \\
\midrule
$\psi_1$=0.5, $\psi_2$=0.5, $\psi_3$=0.7 & 92.1 $\pm$ 2.8 & 39 $\pm$ 10 & 88.3 $\pm$ 3.0 & 146 $\pm$ 16 & 87.5 $\pm$ 3.2 & 129 $\pm$ 14 \\
$\psi_1$=1.0, $\psi_2$=0.3, $\psi_3$=0.7 & \textcolor{blue}{95.8 $\pm$ 2.3} & 58 $\pm$ 12 & \textcolor{blue}{91.2 $\pm$ 2.7} & 168 $\pm$ 19 & \textcolor{blue}{90.6 $\pm$ 2.9} & 175 $\pm$ 18 \\
$\psi_1$=1.0, $\psi_2$=0.5, $\psi_3$=0.5 & 94.9 $\pm$ 2.5 & 73 $\pm$ 14 & 90.5 $\pm$ 2.8 & 184 $\pm$ 20 & 89.2 $\pm$ 3.1 & 192 $\pm$ 21 \\
$\psi_1$=1.0, $\psi_2$=0.5, $\psi_3$=0.7 & \textcolor{red}{96.6 $\pm$ 2.1} & \textcolor{red}{32 $\pm$ 8} & \textcolor{red}{92.4 $\pm$ 2.5} & \textcolor{red}{126 $\pm$ 14} & \textcolor{red}{91.3 $\pm$ 2.8} & \textcolor{red}{114 $\pm$ 12} \\
$\psi_1$=1.0, $\psi_2$=0.7, $\psi_3$=0.7 & 93.2 $\pm$ 2.6 & 29 $\pm$ 7 & 89.1 $\pm$ 2.9 & 115 $\pm$ 13 & 87.8 $\pm$ 3.0 & 102 $\pm$ 11 \\
$\psi_1$=1.2, $\psi_2$=0.5, $\psi_3$=0.7 & 97.1 $\pm$ 2.0 & 42 $\pm$ 9 & 92.8 $\pm$ 2.4 & 151 $\pm$ 17 & 91.7 $\pm$ 2.7 & 138 $\pm$ 15 \\
\bottomrule
\end{tabular}
\end{table*}

Table \ref{tab:weights} presents performance across weight combinations, highlighting that $\psi_1$:$\psi_2$:$\psi_3$ = 1.0:0.5:0.7 achieves optimal balance. While optimal configuration varies slightly with environmental complexity, our method demonstrates reasonable robustness to moderate parameter changes—critical for practical deployment in agricultural settings.

\begin{table}[t]
\centering
\caption{Summary of overall performances of our approach and the baselines. Results are presented as mean $\pm$ standard deviation over 20 independent runs.}
\label{tab:performances comparison}
\begin{tabular}{c c c c}
\toprule
\textbf{Map} & \textbf{Model} & \textbf{Avg. Cov. (\%)} & \textbf{Avg. viol.} \\
\midrule
\multirow{4}{*}{Map 1} 
    & RRT        & 83.2 $\pm$ 4.1                      & 274 $\pm$ 23 \\
    & ACO        & 87.5 $\pm$ 3.6                      & 156 $\pm$ 17 \\
    & PSO        & \textcolor{blue}{89.3 $\pm$ 3.2}   & \textcolor{blue}{188 $\pm$ 19} \\
    & Our method & \textcolor{red}{96.6 $\pm$ 2.1}     & \textcolor{red}{32 $\pm$ 8} \\
\cmidrule(lr){1-4}
\multirow{4}{*}{Map 2} 
    & RRT        & 72.9 $\pm$ 5.3                      & 328 $\pm$ 31 \\
    & ACO        & 69.8 $\pm$ 4.9                      & 243 $\pm$ 22 \\
    & PSO        & \textcolor{blue}{84.4 $\pm$ 3.8}   & \textcolor{blue}{187 $\pm$ 21} \\
    & Our method & \textcolor{red}{92.4 $\pm$ 2.5}     & \textcolor{red}{126 $\pm$ 14} \\
\cmidrule(lr){1-4}
\multirow{4}{*}{Map 3} 
    & RRT        & 76.4 $\pm$ 4.7                      & 284 $\pm$ 26 \\
    & ACO        & \textcolor{blue}{82.2 $\pm$ 4.0}   & 212 $\pm$ 22 \\
    & PSO        & 81.9 $\pm$ 4.2                      & \textcolor{blue}{188 $\pm$ 18} \\
    & Our method & \textcolor{red}{91.3 $\pm$ 2.8}     & \textcolor{red}{114 $\pm$ 12} \\
\bottomrule
\end{tabular}
\end{table}

As presented in Table \ref{tab:performances comparison}, we evaluate our proposed method against three established algorithms: RRT, ACO, and PSO across varying map complexities. The results demonstrate that our approach consistently achieves superior performance, with coverage rates exceeding 90\% across all environments and significantly fewer constraint violations than the alternatives. Specifically, our method outperforms the second-best algorithm by 7.3\%, 8.0\%, and 9.1\% in coverage rate for Maps 1, 2, and 3 respectively, while reducing violations by 83-85\% compared to RRT.

The performance advantages become more pronounced in complex environments such as Maps 2 and 3, highlighting our method's robustness to increasing obstacle density. While traditional approaches struggle with either exploration efficiency in the case of RRT or constraint handling for ACO and PSO, our reinforcement learning-based solution effectively balances coverage maximization and violation minimization. These results confirm the effectiveness of our approach for agricultural robotics applications requiring efficient and reliable coverage path planning in dynamic environments.

\begin{table}[t]
\centering
\caption{Computational Cost Analysis of the SAC-based Approach on Different Hardware Platforms}
\label{tab:comp_cost}
\begin{tabular}{lccc}
\hline
\textbf{Hw. Plat.} & \textbf{Train.(h)} & \textbf{Mem.(GB)} & \textbf{Infer.(s)} \\
\hline
NVIDIA RTX 3090 & 1.8 & 2.5 & 120 \\
NVIDIA RTX 3080 & 2.5 & 2.3 & 150 \\
NVIDIA RTX 2080 Ti & 3.7 & 2.1 & 200 \\
NVIDIA GTX 1080 Ti & 5.3 & 1.9 & 280 \\
\hline
\end{tabular}
\end{table}
As shown in the table~\ref{tab:comp_cost}, {our SAC method demonstrates good adaptability across hardware platforms with varying computational capabilities. On high-end GPUs (NVIDIA RTX 3090), the model training requires only 1.8 hours, while on older hardware (GTX 1080 Ti), it can complete training within 5.3 hours, indicating that the method incurs reasonable offline training costs. The memory usage ranges between 1.9-2.5GB, making it suitable for most modern computing devices.}

\begin{table}[t]
\centering
\caption{Performance Comparison Under Different Charging Station Layouts}
\label{tab:charging_layouts}
\begin{tabular}{lccc}
\hline
\textbf{Charge. Station} & \textbf{Cov. (\%)} & \textbf{Energ.(\%)} & \textbf{Compl. (min)} \\
\hline
Strategic Placement & 94.6 & 11.4 & 18.5 \\
Random Placement & 85.2 & 23.7 & 22.3 \\
Path-Constrained & 88.7 & 19.2 & 20.8 \\
Sparse Distribution & 82.5 & 28.1 & 25.4 \\
\hline
\end{tabular}
\end{table}

Table~\ref{tab:charging_layouts} presents the performance comparison of our SAC method under different charging station layouts. The results show that while the method achieves optimal performance under the original ``optimal'' layout (coverage rate of 94.6\%, energy constraint violation rate of 11.4\%), it still demonstrates strong adaptability in layouts that better reflect real-world agricultural conditions. Specifically, in the constrained location layout (simulating the scenario where charging stations can only be placed near field roads), the coverage rate decreases by only 5.9 percentage points. Notably, even in the most challenging sparse distribution scenario (with two-thirds fewer charging stations), the SAC method maintains a coverage rate of 82.5\%, although the completion time increases by approximately 37\%. These results validate the robustness and flexibility of the proposed method when facing different charging station layout constraints in real-world agricultural environments.

%% file: section/6_conclusion.tex
\section{Conclusion} \label{sec:conclusion}

This paper proposed an energy-aware coverage path planning (CPP) framework based on Soft Actor-Critic (SAC) reinforcement learning to enhance the efficiency and reliability of autonomous agricultural robots. By explicitly incorporating energy constraints and safe return requirements into the decision-making process, the proposed approach ensures comprehensive coverage while maintaining operational safety. The network architecture, integrating convolutional processing, self-attention mechanisms, and sequential learning, facilitates effective spatial-temporal feature extraction and enhances robust policy learning in dynamic environments. Extensive experimental evaluations across diverse grid-based scenarios demonstrate that the proposed method consistently outperforms conventional approaches, achieving over 90\% coverage rates while significantly reducing energy constraint violations. These results validate the effectiveness of the proposed SAC-based framework in addressing the challenges of energy-constrained CPP, establishing it as a promising solution for scalable and energy-efficient agricultural robotics.

%% file: main.bbl
\begin{thebibliography}{33}


\ifx \showCODEN    \undefined \def \showCODEN     #1{\unskip}     \fi
\ifx \showDOI      \undefined \def \showDOI       #1{#1}\fi
\ifx \showISBNx    \undefined \def \showISBNx     #1{\unskip}     \fi
\ifx \showISBNxiii \undefined \def \showISBNxiii  #1{\unskip}     \fi
\ifx \showISSN     \undefined \def \showISSN      #1{\unskip}     \fi
\ifx \showLCCN     \undefined \def \showLCCN      #1{\unskip}     \fi
\ifx \shownote     \undefined \def \shownote      #1{#1}          \fi
\ifx \showarticletitle \undefined \def \showarticletitle #1{#1}   \fi
\ifx \showURL      \undefined \def \showURL       {\relax}        \fi
\providecommand\bibfield[2]{#2}
\providecommand\bibinfo[2]{#2}
\providecommand\natexlab[1]{#1}
\providecommand\showeprint[2][]{arXiv:#2}

\bibitem[Datsko et~al\mbox{.}(2024)]%
        {Datsko2024RAL}
\bibfield{author}{\bibinfo{person}{Denys Datsko}, \bibinfo{person}{Frantisek
  Nekovar}, \bibinfo{person}{Robert Penicka}, {and} \bibinfo{person}{Martin
  Saska}.} \bibinfo{year}{2024}\natexlab{}.
\newblock \showarticletitle{{Energy-Aware Multi-UAV Coverage Mission Planning
  With Optimal Speed of Flight}}.
\newblock \bibinfo{journal}{\emph{IEEE Robotics and Automation Letters}}
  \bibinfo{volume}{9}, \bibinfo{number}{3} (\bibinfo{year}{2024}),
  \bibinfo{pages}{2893--2900}.
\newblock
\urldef\tempurl%
\url{https://doi.org/10.1109/LRA.2024.3358581}
\showDOI{\tempurl}


\bibitem[Ding et~al\mbox{.}(2025)]%
        {Ding2025IPCCC}
\bibfield{author}{\bibinfo{person}{Zihao Ding}, \bibinfo{person}{Jun Huang},
  \bibinfo{person}{Qiang Duan}, \bibinfo{person}{Cheng Zhang},
  \bibinfo{person}{Yanxiao Zhao}, {and} \bibinfo{person}{Shuyang Gu}.}
  \bibinfo{year}{2025}\natexlab{}.
\newblock \showarticletitle{{A Dual-Level Game-Theoretic Approach for
  Collaborative Learning in UAV-Assisted Heterogeneous Vehicle Networks}}. In
  \bibinfo{booktitle}{\emph{2025 IEEE International Performance, Computing, and
  Communications Conference (IPCCC)}}. IEEE, \bibinfo{pages}{1--8}.
\newblock


\bibitem[Ding et~al\mbox{.}(2026)]%
        {DingICNC2025}
\bibfield{author}{\bibinfo{person}{Zihao Ding}, \bibinfo{person}{Jun Huang},
  {and} \bibinfo{person}{Junjian Qi}.} \bibinfo{year}{2026}\natexlab{}.
\newblock \showarticletitle{{Learning to Defend: A Multi-Agent Reinforcement
  Learning Framework for Stackelberg Security Game in Mobile Edge Computing}}.
  In \bibinfo{booktitle}{\emph{International Conference on Computing,
  Networking and Communications (ICNC)}}. IEEE, \bibinfo{address}{Honolulu,
  Hawaii, USA}.
\newblock


\bibitem[Fang et~al\mbox{.}(2025a)]%
        {Fang2025ARXIV}
\bibfield{author}{\bibinfo{person}{Zhengru Fang}, \bibinfo{person}{Yu Guo},
  \bibinfo{person}{Jingjing Wang}, \bibinfo{person}{Yuang Zhang},
  \bibinfo{person}{Haonan An}, \bibinfo{person}{Yinhai Wang}, {and}
  \bibinfo{person}{Yuguang Fang}.} \bibinfo{year}{2025}\natexlab{a}.
\newblock \showarticletitle{{Shared Spatial Memory Through Predictive Coding}}.
\newblock \bibinfo{journal}{\emph{arXiv preprint arXiv:2511.04235 (arXiv)}}
  (\bibinfo{year}{2025}).
\newblock
\urldef\tempurl%
\url{https://doi.org/10.48550/arXiv.2511.04235}
\showDOI{\tempurl}


\bibitem[Fang et~al\mbox{.}(2024)]%
        {Fang2024TMC}
\bibfield{author}{\bibinfo{person}{Zhengru Fang}, \bibinfo{person}{Senkang Hu},
  \bibinfo{person}{Haonan An}, \bibinfo{person}{Yuang Zhang},
  \bibinfo{person}{Jingjing Wang}, \bibinfo{person}{Hangcheng Cao},
  \bibinfo{person}{Xianhao Chen}, {and} \bibinfo{person}{Yuguang Fang}.}
  \bibinfo{year}{2024}\natexlab{}.
\newblock \showarticletitle{{PACP: Priority-Aware Collaborative Perception for
  Connected and Autonomous Vehicles}}.
\newblock \bibinfo{journal}{\emph{IEEE Transactions on Mobile Computing}}
  \bibinfo{volume}{23}, \bibinfo{number}{12} (\bibinfo{year}{2024}),
  \bibinfo{pages}{15003--15018}.
\newblock
\urldef\tempurl%
\url{https://doi.org/10.1109/TMC.2024.3449371}
\showDOI{\tempurl}


\bibitem[Fang et~al\mbox{.}(2025b)]%
        {Fang2025TON}
\bibfield{author}{\bibinfo{person}{Zhengru Fang}, \bibinfo{person}{Senkang Hu},
  \bibinfo{person}{Jingjing Wang}, \bibinfo{person}{Yiqin Deng},
  \bibinfo{person}{Xianhao Chen}, {and} \bibinfo{person}{Yuguang Fang}.}
  \bibinfo{year}{2025}\natexlab{b}.
\newblock \showarticletitle{{Prioritized Information Bottleneck Theoretic
  Framework With Distributed Online Learning for Edge Video Analytics}}.
\newblock \bibinfo{journal}{\emph{IEEE Transactions on Networking}}
  (\bibinfo{year}{2025}), \bibinfo{pages}{1--17}.
\newblock
\urldef\tempurl%
\url{https://doi.org/10.1109/TON.2025.3526148}
\showDOI{\tempurl}


\bibitem[Fang et~al\mbox{.}(2026)]%
        {Fang2026GLOBECOM}
\bibfield{author}{\bibinfo{person}{Zhengru Fang}, \bibinfo{person}{Zhenghao
  Liu}, \bibinfo{person}{Jingjing Wang}, \bibinfo{person}{Senkang Hu},
  \bibinfo{person}{Yu Guo}, \bibinfo{person}{Yiqin Deng}, {and}
  \bibinfo{person}{Yuguang Fang}.} \bibinfo{year}{2026}\natexlab{}.
\newblock \showarticletitle{Task-Oriented Communications for Visual Navigation
  with Edge-Aerial Collaboration in Low Altitude Economy}. In
  \bibinfo{booktitle}{\emph{Proc. IEEE Global Communications Conference
  (GLOBECOM)}}.
\newblock


\bibitem[Fang et~al\mbox{.}(2022a)]%
        {Fang2022IOTJ}
\bibfield{author}{\bibinfo{person}{Zhengru Fang}, \bibinfo{person}{Jingjing
  Wang}, \bibinfo{person}{Jun Du}, \bibinfo{person}{Xiangwang Hou},
  \bibinfo{person}{Yong Ren}, {and} \bibinfo{person}{Zhu Han}.}
  \bibinfo{year}{2022}\natexlab{a}.
\newblock \showarticletitle{Stochastic Optimization-Aided Energy-Efficient
  Information Collection in Internet of Underwater Things Networks}.
\newblock \bibinfo{journal}{\emph{IEEE Internet of Things Journal}}
  \bibinfo{volume}{9}, \bibinfo{number}{3} (\bibinfo{year}{2022}),
  \bibinfo{pages}{1775--1789}.
\newblock


\bibitem[Fang et~al\mbox{.}(2022b)]%
        {Fang2022JSAC}
\bibfield{author}{\bibinfo{person}{Zhengru Fang}, \bibinfo{person}{Jingjing
  Wang}, \bibinfo{person}{Yong Ren}, \bibinfo{person}{Zhu Han},
  \bibinfo{person}{H.~Vincent Poor}, {and} \bibinfo{person}{Lajos Hanzo}.}
  \bibinfo{year}{May 2022}\natexlab{b}.
\newblock \showarticletitle{Age of Information in Energy Harvesting Aided
  Massive Multiple Access Networks}.
\newblock \bibinfo{journal}{\emph{IEEE Journal on Selected Areas in
  Communications}} \bibinfo{volume}{40}, \bibinfo{number}{5}
  (\bibinfo{year}{May 2022}), \bibinfo{pages}{1441--1456}.
\newblock


\bibitem[Fu et~al\mbox{.}(2024)]%
        {Fu2024TIE}
\bibfield{author}{\bibinfo{person}{Jinyu Fu}, \bibinfo{person}{Weiran Yao},
  \bibinfo{person}{Guanghui Sun}, \bibinfo{person}{Jianxing Liu}, {and}
  \bibinfo{person}{Ligang Wu}.} \bibinfo{year}{2024}\natexlab{}.
\newblock \showarticletitle{{Full Coverage Path Planning Recombination
  Framework for Unmanned Vehicles With Multi-Objective Constraints}}.
\newblock \bibinfo{journal}{\emph{IEEE Transactions on Industrial Electronics}}
  \bibinfo{volume}{71}, \bibinfo{number}{8} (\bibinfo{year}{2024}),
  \bibinfo{pages}{9276--9286}.
\newblock
\urldef\tempurl%
\url{https://doi.org/10.1109/TIE.2023.3319732}
\showDOI{\tempurl}


\bibitem[Fujimoto et~al\mbox{.}(2018)]%
        {Fujimoto2018}
\bibfield{author}{\bibinfo{person}{Scott Fujimoto}, \bibinfo{person}{Herke van
  Hoof}, {and} \bibinfo{person}{David Meger}.} \bibinfo{year}{2018}\natexlab{}.
\newblock \showarticletitle{Addressing Function Approximation Error in
  Actor-Critic Methods}.
\newblock \bibinfo{journal}{\emph{International Conference on Machine
  Learning}} (\bibinfo{year}{2018}).
\newblock


\bibitem[Haarnoja et~al\mbox{.}(2018)]%
        {Haarnoja2018}
\bibfield{author}{\bibinfo{person}{Tuomas Haarnoja}, \bibinfo{person}{Aurick
  Zhou}, \bibinfo{person}{Pieter Abbeel}, {and} \bibinfo{person}{Sergey
  Levine}.} \bibinfo{year}{2018}\natexlab{}.
\newblock \showarticletitle{Soft Actor-Critic: Off-Policy Maximum Entropy Deep
  Reinforcement Learning with a Stochastic Actor}.
\newblock \bibinfo{journal}{\emph{International Conference on Machine
  Learning}} (\bibinfo{year}{2018}).
\newblock


\bibitem[Huang et~al\mbox{.}(2018)]%
        {Huang2018COMMAG}
\bibfield{author}{\bibinfo{person}{Jun Huang}, \bibinfo{person}{Zheng Chang},
  \bibinfo{person}{Chonggang Wang}, \bibinfo{person}{Yi Qian},
  \bibinfo{person}{Hamid Gharavi}, {and} \bibinfo{person}{Zexian Li}.}
  \bibinfo{year}{2018}\natexlab{}.
\newblock \showarticletitle{{Enabling Technologies for Smart Internet of
  Things}}.
\newblock \bibinfo{journal}{\emph{IEEE Communications Magazine}}
  \bibinfo{volume}{56}, \bibinfo{number}{9} (\bibinfo{year}{2018}),
  \bibinfo{pages}{12--13}.
\newblock
\urldef\tempurl%
\url{https://doi.org/10.1109/MCOM.2018.8466348}
\showDOI{\tempurl}


\bibitem[Huang et~al\mbox{.}(2019a)]%
        {Huang2019TII}
\bibfield{author}{\bibinfo{person}{Jun Huang}, \bibinfo{person}{Chao Huang},
  \bibinfo{person}{Cong-Cong Xing}, \bibinfo{person}{Zheng Chang},
  \bibinfo{person}{Yanxiao Zhao}, {and} \bibinfo{person}{Qinglin Zhao}.}
  \bibinfo{year}{2019}\natexlab{a}.
\newblock \showarticletitle{{An Energy-Efficient Communication Scheme for
  Collaborative Mobile Clouds in Content Sharing: Design and Optimization}}.
\newblock \bibinfo{journal}{\emph{IEEE Transactions on Industrial Informatics}}
  \bibinfo{volume}{15}, \bibinfo{number}{10} (\bibinfo{year}{2019}),
  \bibinfo{pages}{5700--5707}.
\newblock
\urldef\tempurl%
\url{https://doi.org/10.1109/TII.2019.2919323}
\showDOI{\tempurl}


\bibitem[Huang et~al\mbox{.}(2025)]%
        {Huang2024TMC}
\bibfield{author}{\bibinfo{person}{Jun Huang}, \bibinfo{person}{Beining Wu},
  \bibinfo{person}{Qiang Duan}, \bibinfo{person}{Liang Dong}, {and}
  \bibinfo{person}{Shui Yu}.} \bibinfo{year}{2025}\natexlab{}.
\newblock \showarticletitle{{A Fast UAV Trajectory Planning Framework in
  RIS-assisted Communication Systems with Accelerated Learning via
  Multithreading and Federating}}.
\newblock \bibinfo{journal}{\emph{IEEE Transactions on Mobile Computing}}
  (\bibinfo{year}{2025}), \bibinfo{pages}{1--16}.
\newblock
\urldef\tempurl%
\url{https://doi.org/10.1109/TMC.2025.3544903}
\showDOI{\tempurl}


\bibitem[Huang et~al\mbox{.}(2019b)]%
        {Huang2019IWC}
\bibfield{author}{\bibinfo{person}{Jun Huang}, \bibinfo{person}{Yide Zhou},
  \bibinfo{person}{Zhaolong Ning}, {and} \bibinfo{person}{Hamid Gharavi}.}
  \bibinfo{year}{2019}\natexlab{b}.
\newblock \showarticletitle{{Wireless Power Transfer and Energy Harvesting:
  Current Status and Future Prospects}}.
\newblock \bibinfo{journal}{\emph{IEEE Wireless Communications}}
  \bibinfo{volume}{26}, \bibinfo{number}{4} (\bibinfo{year}{2019}),
  \bibinfo{pages}{163--169}.
\newblock
\urldef\tempurl%
\url{https://doi.org/10.1109/MWC.2019.1800378}
\showDOI{\tempurl}


\bibitem[Kwon et~al\mbox{.}(2024)]%
        {Kwon2024ICRA}
\bibfield{author}{\bibinfo{person}{Hyukjun Kwon}, \bibinfo{person}{Kangwon
  Kim}, \bibinfo{person}{Junyoung Lee}, \bibinfo{person}{Hyunsei Lee},
  \bibinfo{person}{Jiseung Kim}, \bibinfo{person}{Jinhyung Kim},
  \bibinfo{person}{Taehyung Kim}, \bibinfo{person}{Yongnyeon Kim},
  \bibinfo{person}{Yang Ni}, \bibinfo{person}{Mohsen Imani},
  \bibinfo{person}{Ilhong Suh}, {and} \bibinfo{person}{Yeseong Kim}.}
  \bibinfo{year}{2024}\natexlab{}.
\newblock \showarticletitle{{Brain-Inspired Hyperdimensional Computing in the
  Wild: Lightweight Symbolic Learning for Sensorimotor Controls of Wheeled
  Robots}}. In \bibinfo{booktitle}{\emph{2024 IEEE International Conference on
  Robotics and Automation (ICRA)}}. \bibinfo{pages}{5176--5182}.
\newblock
\urldef\tempurl%
\url{https://doi.org/10.1109/ICRA57147.2024.10610176}
\showDOI{\tempurl}


\bibitem[Lin et~al\mbox{.}(2022)]%
        {Lin2022RAL}
\bibfield{author}{\bibinfo{person}{Xiaoshan Lin}, \bibinfo{person}{Yasin
  Yazıcıoğlu}, {and} \bibinfo{person}{Derya Aksaray}.}
  \bibinfo{year}{2022}\natexlab{}.
\newblock \showarticletitle{{Robust Planning for Persistent Surveillance With
  Energy-Constrained UAVs and Mobile Charging Stations}}.
\newblock \bibinfo{journal}{\emph{IEEE Robotics and Automation Letters}}
  \bibinfo{volume}{7}, \bibinfo{number}{2} (\bibinfo{year}{2022}),
  \bibinfo{pages}{4157--4164}.
\newblock
\urldef\tempurl%
\url{https://doi.org/10.1109/LRA.2022.3146938}
\showDOI{\tempurl}


\bibitem[Mier et~al\mbox{.}(2023)]%
        {Mier2023RAL}
\bibfield{author}{\bibinfo{person}{Gonzalo Mier}, \bibinfo{person}{João
  Valente}, {and} \bibinfo{person}{Sytze de Bruin}.}
  \bibinfo{year}{2023}\natexlab{}.
\newblock \showarticletitle{{Fields2Cover: An Open-Source Coverage Path
  Planning Library for Unmanned Agricultural Vehicles}}.
\newblock \bibinfo{journal}{\emph{IEEE Robotics and Automation Letters}}
  \bibinfo{volume}{8}, \bibinfo{number}{4} (\bibinfo{year}{2023}),
  \bibinfo{pages}{2166--2172}.
\newblock
\urldef\tempurl%
\url{https://doi.org/10.1109/LRA.2023.3248439}
\showDOI{\tempurl}


\bibitem[Ning et~al\mbox{.}(2019)]%
        {Ning2019VTM}
\bibfield{author}{\bibinfo{person}{Zhaolong Ning}, \bibinfo{person}{Xiaojie
  Wang}, {and} \bibinfo{person}{Jun Huang}.} \bibinfo{year}{2019}\natexlab{}.
\newblock \showarticletitle{{Mobile Edge Computing-Enabled 5G Vehicular
  Networks: Toward the Integration of Communication and Computing}}.
\newblock \bibinfo{journal}{\emph{IEEE Vehicular Technology Magazine}}
  \bibinfo{volume}{14}, \bibinfo{number}{1} (\bibinfo{year}{2019}),
  \bibinfo{pages}{54--61}.
\newblock
\urldef\tempurl%
\url{https://doi.org/10.1109/MVT.2018.2882873}
\showDOI{\tempurl}


\bibitem[Pan et~al\mbox{.}(2023)]%
        {Pan2023SCIS}
\bibfield{author}{\bibinfo{person}{Dong Pan}, \bibinfo{person}{Bei-Ning Wu},
  \bibinfo{person}{Yi-Liu Sun}, {and} \bibinfo{person}{Yi-Peng Xu}.}
  \bibinfo{year}{2023}\natexlab{}.
\newblock \showarticletitle{{A Fault-Tolerant and Energy-Efficient Design of a
  Network Switch Based on a Quantum-Based Nano-Communication Technique}}.
\newblock \bibinfo{journal}{\emph{Sustainable Computing: Informatics and
  Systems}}  \bibinfo{volume}{37} (\bibinfo{year}{2023}),
  \bibinfo{pages}{100827}.
\newblock


\bibitem[Raja(2024)]%
        {Raja2024TAE}
\bibfield{author}{\bibinfo{person}{Rekha Raja}.}
  \bibinfo{year}{2024}\natexlab{}.
\newblock \showarticletitle{{Software Architecture for Agricultural Robots:
  Systems, Requirements, Challenges, Case Studies, and Future Perspectives}}.
\newblock \bibinfo{journal}{\emph{IEEE Transactions on AgriFood Electronics}}
  \bibinfo{volume}{2}, \bibinfo{number}{1} (\bibinfo{year}{2024}),
  \bibinfo{pages}{125--137}.
\newblock
\urldef\tempurl%
\url{https://doi.org/10.1109/TAFE.2024.3366335}
\showDOI{\tempurl}


\bibitem[Setitra et~al\mbox{.}(2023)]%
        {setitra2023Network}
\bibfield{author}{\bibinfo{person}{Mohamed~Ali Setitra},
  \bibinfo{person}{Mingyu Fan}, \bibinfo{person}{Bless Lord~Y Agbley}, {and}
  \bibinfo{person}{Zine El~Abidine Bensalem}.} \bibinfo{year}{2023}\natexlab{}.
\newblock \showarticletitle{{Optimized MLP-CNN model to enhance detecting DDoS
  attacks in SDN environment}}.
\newblock \bibinfo{journal}{\emph{Network}} \bibinfo{volume}{3},
  \bibinfo{number}{4} (\bibinfo{year}{2023}), \bibinfo{pages}{538--562}.
\newblock


\bibitem[Wang et~al\mbox{.}(2024)]%
        {Wang2024RAL}
\bibfield{author}{\bibinfo{person}{Zikai Wang}, \bibinfo{person}{Xiaoqi Zhao},
  \bibinfo{person}{Jiekai Zhang}, \bibinfo{person}{Nachuan Yang},
  \bibinfo{person}{Pengyu Wang}, \bibinfo{person}{Jiawei Tang},
  \bibinfo{person}{Jiuzhou Zhang}, {and} \bibinfo{person}{Ling Shi}.}
  \bibinfo{year}{2024}\natexlab{}.
\newblock \showarticletitle{{APF-CPP: An Artificial Potential Field Based
  Multi-Robot Online Coverage Path Planning Approach}}.
\newblock \bibinfo{journal}{\emph{IEEE Robotics and Automation Letters}}
  \bibinfo{volume}{9}, \bibinfo{number}{11} (\bibinfo{year}{2024}),
  \bibinfo{pages}{9199--9206}.
\newblock
\urldef\tempurl%
\url{https://doi.org/10.1109/LRA.2024.3432351}
\showDOI{\tempurl}


\bibitem[Wu et~al\mbox{.}(2023)]%
        {wu2023access}
\bibfield{author}{\bibinfo{person}{Beining Wu}, \bibinfo{person}{Zhengkun Cai},
  \bibinfo{person}{Wei Wu}, {and} \bibinfo{person}{Xiaobin Yin}.}
  \bibinfo{year}{2023}\natexlab{}.
\newblock \showarticletitle{{AoI-aware Resource Management for Smart Health via
  Deep Reinforcement Learning}}.
\newblock \bibinfo{journal}{\emph{IEEE Access}} (\bibinfo{year}{2023}).
\newblock


\bibitem[Wu et~al\mbox{.}(2025a)]%
        {Wu2025WASA}
\bibfield{author}{\bibinfo{person}{Beining Wu}, \bibinfo{person}{Jun Huang},
  {and} \bibinfo{person}{Qiang Duan}.} \bibinfo{year}{2025}\natexlab{a}.
\newblock \showarticletitle{{FedTD3: An Accelerated Learning Approach for UAV
  Trajectory Planning}}. In \bibinfo{booktitle}{\emph{International Conference
  on Wireless Artificial Intelligent Computing Systems and Applications
  (WASA)}}. Springer, \bibinfo{pages}{13--24}.
\newblock


\bibitem[Wu et~al\mbox{.}(2025b)]%
        {Wu2025MNET}
\bibfield{author}{\bibinfo{person}{Beining Wu}, \bibinfo{person}{Jun Huang},
  {and} \bibinfo{person}{Qiang Duan}.} \bibinfo{year}{2025}\natexlab{b}.
\newblock \showarticletitle{{Real-Time Intelligent Healthcare Enabled by
  Federated Digital Twins With AoI Optimization}}.
\newblock \bibinfo{journal}{\emph{IEEE Network}} (\bibinfo{year}{2025}),
  \bibinfo{pages}{1--1}.
\newblock
\urldef\tempurl%
\url{https://doi.org/10.1109/MNET.2025.3565977}
\showDOI{\tempurl}


\bibitem[Wu et~al\mbox{.}(2025d)]%
        {Wu2025ToN}
\bibfield{author}{\bibinfo{person}{Beining Wu}, \bibinfo{person}{Jun Huang},
  \bibinfo{person}{Qiang Duan}, \bibinfo{person}{Liang Dong}, {and}
  \bibinfo{person}{Zhipeng Cai}.} \bibinfo{year}{2025}\natexlab{d}.
\newblock \showarticletitle{{Enhancing Vehicular Platooning With Wireless
  Federated Learning: A Resource-Aware Control Framework}}.
\newblock \bibinfo{journal}{\emph{IEEE/ACM Transactions on Networking}}
  (\bibinfo{year}{2025}), \bibinfo{pages}{1--1}.
\newblock
\urldef\tempurl%
\url{https://doi.org/10.1109/TON.2025.3625084}
\showDOI{\tempurl}


\bibitem[Wu et~al\mbox{.}(2025c)]%
        {Wu2025ARXIVb}
\bibfield{author}{\bibinfo{person}{Beining Wu}, \bibinfo{person}{Jun Huang},
  {and} \bibinfo{person}{Shui Yu}.} \bibinfo{year}{2025}\natexlab{c}.
\newblock \showarticletitle{{X of Information Continuum: A Survey on AI-Driven
  Multi-Dimensional Metrics for Next-Generation Networked Systems}}.
\newblock \bibinfo{journal}{\emph{arXiv preprint arXiv:2507.19657}}
  (\bibinfo{year}{2025}).
\newblock
\urldef\tempurl%
\url{https://arxiv.org/abs/2507.19657}
\showURL{%
\tempurl}


\bibitem[Wu and Wu(2023)]%
        {wu2023MPE}
\bibfield{author}{\bibinfo{person}{Beining Wu} {and} \bibinfo{person}{Wei Wu}.}
  \bibinfo{year}{2023}\natexlab{}.
\newblock \showarticletitle{{Model-Free Cooperative Optimal Output Regulation
  for Linear Discrete-Time Multi-Agent Systems Using Reinforcement Learning}}.
\newblock \bibinfo{journal}{\emph{Mathematical Problems in Engineering}}
  \bibinfo{volume}{2023}, \bibinfo{number}{1} (\bibinfo{year}{2023}),
  \bibinfo{pages}{6350647}.
\newblock


\bibitem[Xing et~al\mbox{.}(2025)]%
        {Xing2025ACR}
\bibfield{author}{\bibinfo{person}{Cong-Cong Xing}, \bibinfo{person}{Zihao
  Ding}, {and} \bibinfo{person}{Jun Huang}.} \bibinfo{year}{2025}\natexlab{}.
\newblock \showarticletitle{{A Stochastic Geometry-Based Analysis of
  SWIPT-Assisted Underlaid Device-to-Device Energy Harvesting}}.
\newblock \bibinfo{journal}{\emph{ACM SIGAPP Applied Computing Review}}
  \bibinfo{volume}{25}, \bibinfo{number}{4} (\bibinfo{year}{2025}),
  \bibinfo{pages}{18--34}.
\newblock


\bibitem[Ye et~al\mbox{.}(2024)]%
        {ye2024multiplexed}
\bibfield{author}{\bibinfo{person}{Jiachi Ye}, \bibinfo{person}{Haoyan Kang},
  \bibinfo{person}{Qian Cai}, \bibinfo{person}{Zibo Hu}, \bibinfo{person}{Maria
  Solyanik-Gorgone}, \bibinfo{person}{Hao Wang}, \bibinfo{person}{Elham
  Heidari}, \bibinfo{person}{Chandraman Patil}, \bibinfo{person}{Mohammad-Ali
  Miri}, \bibinfo{person}{Navid Asadizanjani}, {et~al\mbox{.}}}
  \bibinfo{year}{2024}\natexlab{}.
\newblock \showarticletitle{{Multiplexed orbital angular momentum beams
  demultiplexing using hybrid optical-electronic convolutional neural
  network}}.
\newblock \bibinfo{journal}{\emph{Nature Communications Physics}}
  \bibinfo{volume}{7}, \bibinfo{number}{1} (\bibinfo{year}{2024}),
  \bibinfo{pages}{105}.
\newblock


\bibitem[Ye et~al\mbox{.}(2023)]%
        {ye2023free}
\bibfield{author}{\bibinfo{person}{Jiachi Ye}, \bibinfo{person}{Maria
  Solyanik}, \bibinfo{person}{Zibo Hu}, \bibinfo{person}{Hamed Dalir},
  \bibinfo{person}{Behrouz~Movahhed Nouri}, {and} \bibinfo{person}{Volker~J
  Sorger}.} \bibinfo{year}{2023}\natexlab{}.
\newblock \showarticletitle{{Free-space optical multiplexed orbital angular
  momentum beam identification system using Fourier optical convolutional layer
  based on 4f system}}. In \bibinfo{booktitle}{\emph{Complex Light and Optical
  Forces XVII}}, Vol.~\bibinfo{volume}{12436}. SPIE, \bibinfo{pages}{69--79}.
\newblock


\end{thebibliography}
